\pdfoutput=1 %
\documentclass{article} %
\usepackage{preprint,times}

\usepackage{amsmath,amsfonts,bm}

\def\eqref#1{equation~\ref{#1}}

\def\1{\bm{1}}

\DeclareMathAlphabet{\mathsfit}{\encodingdefault}{\sfdefault}{m}{sl}
\SetMathAlphabet{\mathsfit}{bold}{\encodingdefault}{\sfdefault}{bx}{n}

\usepackage{hyperref}
\usepackage{url}
\usepackage{booktabs}
\usepackage{multirow}
\usepackage{cleveref}

\usepackage{graphicx}

\usepackage{algorithm}
\usepackage{algpseudocode}

\renewcommand{\paragraph}[1]{%
  \par\vspace{0.25em}%
  \noindent\textbf{#1}
}

\headertext{}

\title{Rethinking MoE Load Balancing as \\ Feedback Control}

\author{Yikun Hou \& Alp Yurtsever \\
Department of Mathematics and Mathematical Statistics \\
Umeå University, Sweden \\
\texttt{\{yikun.hou,alp.yurtsever\}@umu.se}
}

\begin{document}

\maketitle

\begin{abstract}
Mixture-of-Experts (MoE) training is often not stationary: the data mixture changes at curriculum boundaries, when a model is continually pretrained into a new domain, or when the corpus is refreshed.
Routing is learned against the current mixture, so each such change disrupts the per-expert load the router had converged to, and existing balancing methods have only been evaluated on stationary data.
We rethink load balancing as a feedback control problem, in which the deviation of each expert's load from its fair share is the control error and the per-expert routing bias is the control input, and define a family of controllers, the PID-Balancer, from the proportional, integral, and derivative terms of classical PID control.
Two widely used methods belong to this family: the auxiliary balancing loss behaves like a proportional controller, and the loss-free bias update is an integral controller acting on the sign of the error.
Reading the router as a static map with a one-step delay, we show that the integral term is the only one that both removes persistent imbalance and averages out the step-to-step fluctuation of the measured load, whereas the proportional and derivative terms pass or amplify that fluctuation into the routing bias.
We therefore adopt PID (I), the pure-integral member of the family with a single gain. 
Two settings are evaluated: (i) training an OLMoE backbone from scratch with 16 to 64 experts per layer, through a mid-run mixture switch, and (ii) continual pretraining of the released OLMoE-1B-7B on a new corpus.
PID (I) holds balance before the shift and restores it more tightly afterwards than either the auxiliary loss or loss-free balancing, at comparable perplexity.
\end{abstract}

\section{Introduction}
\label{sec:introduction}

Mixture-of-Experts (MoE) architectures decouple model capacity from per-token compute by replacing dense feed-forward layers with sparsely-activated expert sub-networks \citep{fedus2022switch, dai2024deepseekmoe, liu2024deepseek}.  
Realizing this promise depends critically on one constraint: tokens must be distributed across experts rather than collapsing onto a small subset. 
Without active intervention, MoE routers exhibit a rich-get-richer dynamic. Early-trained experts attract more routing than others, gradient signal concentrates on them, and the remaining experts receive less training and become weaker \citep{Eigen2013LearningFR, shazeer2017}.

Load balancing is the mechanism that prevents this collapse, and existing methods fall into two families.
The first adds an auxiliary loss that penalizes uneven expert usage, weighted by a tunable coefficient.
This balancing pressure enters the router through the same backward pass as the language-modeling objective, and the two combine into a single gradient update to the router weights.
The second, introduced by \citet{wang2024auxiliary} and popularized through its deployment in DeepSeek-V3 \citep{liu2024deepseek}, moves that pressure out of the gradient.
A per-expert bias, updated at each step in proportion to the \emph{sign} of the load deviation, shifts which experts are selected while leaving the router weights to the language-modeling objective alone.
\citet{han2025theoretical} formalize this update as a single-shot primal-dual step for an assignment problem, in which the per-expert bias is identified with the dual variable of the load constraint.

Both families were largely developed and evaluated under a stationary training distribution, in which the per-expert loads tend to drift slowly and a balancing mechanism can settle and then hold.
However, training is not always stationary.
The distribution changes at a curriculum stage boundary, when continual pretraining extends a model into a new domain, or when the corpus itself is refreshed, and each such change disrupts the routing the model had converged to.
We study load balancing in this setting along three axes, which are whether a mechanism maintains balance before the change, whether it restores balance after the change, and how much task performance it sacrifices.

We formulate load balancing as a feedback control loop.
The imbalance of each expert, defined as its deviation from an equal share of the batch, is the control error, and a per-expert bias added to the router scores at selection time is the control input.
Based on this formulation, we propose PID-Balancer, a discrete-time controller whose output combines the current error, its accumulated history, and its most recent change through three gains $(K_P, K_I, K_D)$.
By analyzing and ablating all three terms, we find that the integral term is the only one that both drives persistent imbalance to zero and prevents per-step fluctuations in the measured load from entering the bias.
We therefore deploy PID (I), a linear integral controller with a single gain $K_I$, as our balancer.
It achieves better load balance than all compared baselines in both the pretraining and continual pretraining settings.

Contributions are summarized as follows: 
\vspace{0.1em}
\newline
\indent\hspace{1.2em} \textbf{A unified control framework for MoE load balancing.}
We formulate the routing as a feedback loop, with the per-expert imbalance as the error and the bias added to the router scores as the control input.
Within this formulation, we define PID-Balancer and provide a unified view of two widely used methods.
ALF-LB \citep{wang2024auxiliary} is exactly a relay integrator, and the auxiliary loss \citep{fedus2022switch} acts as a proportional mechanism.
\vspace{0.1em}
\newline
\indent\hspace{1.2em} \textbf{A principled integral balancer.}
Under the control formulation, the map from bias to load is static and monotone, with a one-step delay, and it forms a feedback loop.
For such a loop, only the integral term removes persistent imbalance and averages out per-step load fluctuations through its memory of past errors.
We therefore adopt the integral-only PID (I) as our default balancer.
\vspace{0.1em}
\newline
\indent\hspace{1.2em} \textbf{A drop-in balancer for pretrained models. }
PID (I) modifies neither the model weights nor the training objective, and it can be inserted directly into an existing pretrained model.
We evaluate it on OLMoE-based models in both pretraining and continual pretraining, where PID (I) achieves the best load balance in every setting while maintaining comparable perplexity.

\section{Related Work}
\label{sec:relatedwork}

\paragraph{Load balancing.}
The MoE paradigm \citep{shazeer2017} reduced per-token compute by routing each token to a sparse subset of experts.
Without intervention, routing is self-reinforcing because experts that draw slightly more tokens early in training receive more gradient updates and become better suited to those tokens, which attracts even more tokens \citep{Eigen2013LearningFR,shazeer2017}.
Computation then concentrates on a shrinking subset of experts while the rest remain under-trained.
A range of approaches has been proposed to mitigate this problem, including auxiliary losses with capacity limits \citep{lepikhin2021gshard}, balanced token-to-expert assignment \citep{lewis2021base}, and expert-choice routing \citep{zhou2022mixture}.
Switch Transformer \citep{fedus2022switch} introduced a soft load-balancing loss that has since become one of the dominant approaches.
The DeepSeek series \citep{dai2024deepseekmoe, liu2024deepseek} adopted auxiliary-loss-free balancing \citep{wang2024auxiliary}, showing that gradient-free corrective bias can match or exceed loss-based balancing.
\citet{team2025longcat} adopted a reformulated variant of this approach in LongCat MoE.
\citet{huang2026learning} report that the auxiliary-loss-free formulation alone still results in unstable load balancing in practice.
A separate line of work addresses imbalance at inference time by enforcing capacity limits on a frozen model to bound the latency of overloaded experts, see \citep{he2026capacity} and references therein.

\paragraph{Distribution shift.}
Distribution shift is common in MoE training and can arise in several forms.
(i) Continual pretraining \citep{gururangan2020dont} extends a base model into a new domain, as DeepSeek-Coder-V2 does when it continues pretraining on a code-heavy mixture \citep{zhu2024deepseek}.
A related line of work combines continual pretraining with architecture conversion, as in Innovator, which converts Qwen2.5-7B into a fine-grained scientific MoE during continual pretraining \citep{liao2025innovator}.
\citet{therien2025continual} study how robust MoE routers are to the shift in this setting.
(ii) A curriculum boundary shifts the data mixture during training, as in the three-phase pretraining curriculum of Mellum 2 \citep{kojic2026mellum2}.
This practice is also common in dense-model training, for example in the late-stage annealing mix of OLMo 2 \citep{walsh2025}.
(iii) Supervised fine-tuning \citep{ouyang2022training} narrows the distribution to a single downstream task, and \citet{joseph2026mixture} show that pre-existing routing collapse persists through fine-tuning.
From a control perspective, each form presents the router with a persistent new imbalance after it has already settled once.

\paragraph{Proportional-Integral-Derivative (PID) control.}
Feedback control regulates a system by continuously measuring its output, comparing it against a desired setpoint, and correcting the resulting error.
This paradigm has been developed and formalized over more than a century of control theory \citep{astrom2008feedback}.
The PID controller remains the most widely deployed feedback controller in industrial automation \citep{astrom1995pid}.
It combines the current error, its accumulation, and its rate of change into a single correction.
Within machine learning, PID ideas have been used to accelerate optimization \citep{wang2020pid, shi2020rethinking}, balance loss terms in generative models \citep{shao2020controlvae}, and stabilize constrained reinforcement learning \citep{stooke2020responsive}.

\section{Methodology}
\label{sec:methodlogy}

We rethink load balancing as an instance of feedback control. Under this view, the balancer requires a measured error, an actuation input, and a router that maps the input back to the measured quantity.

\paragraph{Error.}
Perfect balance indicates $T$ tokens each activate $K$ of the $E$ experts, with every expert receiving an equal share $r = TK/E$.
Measuring each expert's deviation from that ideal gives
\begin{equation}
\label{equ:Error}
    e_i(t) = \frac{r - L_i(t)}{r}
\end{equation}
where $L_i(t)$\footnote{Layer index is omitted throughout since the balancer operates independently within each MoE layer.} is the load of expert $i$ at optimizer step $t$.
A positive $e_i(t)$ marks an under-loaded expert, and a negative $e_i(t)$ marks an over-loaded one.
Normalizing by $r$ makes the error dimensionless and bounds it to $[1 - E/K, 1]$ regardless of batch size.

\paragraph{Input.}
The actuation input is a per-expert bias $u_i(t)$ that is computed from $e(t)$ and applied with a \emph{one-step delay}.
The router at step $t$ therefore selects
\begin{equation}
\label{equ:Gate}
\mathcal{T}_j(t) = \operatorname{Top}\text{-}K_i\big(s_{ij}(t) + u_i(t-1)\big),
\qquad
s_{ij}(t) = \frac{\exp z_{ij}(t)}{\sum_{k=1}^{E} \exp z_{kj}(t)},
\end{equation}
where $z_{ij}(t)$ is the router logit of token $j$ for expert $i$, $s_{ij}(t)$ is the corresponding softmax score, and $\mathcal{T}_j(t)$ is the set of $K$ experts selected for token $j$.
The load in \Cref{equ:Error} counts the tokens that select expert $i$, $L_i = \sum_{j=1}^{T} \mathbf{1}\,[\, i \in \mathcal{T}_j \,]$.
The bias is shared by all tokens within an optimizer step, and adjusting $u_i(t-1)$ shifts tokens toward or away from expert $i$ at step $t$.
The bias affects only expert selection because the gate weight remains the unbiased score $s_{ij}(t)$.
It is updated only by the balancer and receives no gradient.

\paragraph{Router and loop.}
The router of \Cref{equ:Gate} maps the bias vector to the load vector. 
Together with the measurement in \Cref{equ:Error} and the balancer of \Cref{subsec:balancer}, it forms a closed loop
\begin{equation}
\label{equ:Loop}
u(t-1)
\;\xrightarrow{\;\text{router}\;}\;
L(t)
\;\xrightarrow{\;\text{measurement}\;}\;
e(t)
\;\xrightarrow{\;\text{balancer}\;}\;
u(t)
\;\to\;\cdots
\end{equation}
The router is \emph{memoryless in the bias}: $L(t)$ depends on the bias $u(t-1)$, given the current scores $s_{ij}(t)$.
The router is \emph{monotone}: raising $u_i$ with the other biases fixed can only enlarge the set of tokens for which expert $i$ enters the $\operatorname{Top}\text{-}K$.
Holding the scores fixed, these properties and the one-step delay place all memory acting on the bias in the balancer, which allows the local analysis in \Cref{subsec:terms} without explicitly modeling the router.

\subsection{The PID-Balancer}
\label{subsec:balancer}

Traditional PID control computes its output as a weighted combination of the present error, the accumulated error, and the error's rate of change.
We work with its discrete-time form and apply it to the error of \Cref{equ:Error}:
\begin{equation}
\label{equ:DiscretePID}
    u_i(t) = K_P\, e_i(t) + K_I \sum_{\tau=0}^{t} e_i(\tau) + K_D\big(e_i(t) - e_i(t-1)\big).
\end{equation}
The three terms read the same error on different time scales.
The proportional term acts on the error at the current step, the integral term accumulates all past error and is the only one with memory, and the derivative term responds to the error's last change.
\Cref{equ:DiscretePID} gives one balancer per expert in each MoE layer, all sharing a single set of gains $(K_P, K_I, K_D)$.
Each balancer output shifts expert selection through \Cref{equ:Gate}.
We call this family the PID-Balancer and provide pseudocode in \Cref{app:pseudocode}.
Each member is named by the terms it keeps.
For example, the pure-integral member PID (I) sets $K_P = K_D = 0$.

\subsection{Existing methods as members of the family}
\label{subsec:auxandalflb}

\paragraph{Auxiliary loss \citep{fedus2022switch} (Aux-loss) acts as a \textit{Proportional} mechanism.}
Aux-loss adds $\mathcal{L}_{\mathrm{aux}} = \alpha E \sum_{i} f_i(t) P_i(t)$ to the training objective, where $f_i(t) = L_i(t)/(TK)$ is the fraction of routing slots taken by expert $i$ and $P_i(t) = \frac{1}{T}\sum_j s_{ij}(t)$ is its mean softmax score over the batch.
Only $P_i(t)$ is differentiable, and $E f_i(t) = 1 - e_i(t)$ by \Cref{equ:Error}, so the gradient on the score of token $j$ for expert $i$ is $\partial \mathcal{L}_{\mathrm{aux}} / \partial s_{ij}(t) = \frac{\alpha}{T}\big(1 - e_i(t)\big)$.
The Jacobian of the softmax is $\partial s_{ij}(t) / \partial z_{kj}(t) = s_{ij}(t)\big(\delta_{ik} - s_{kj}(t)\big)$, where $\delta_{ik} = 1$ if $i = k$ and $0$ otherwise.
Applying the chain rule on the logit $z_{kj}(t)$ of token $j$ for expert $k$ gives
\begin{equation}
\label{equ:AuxChain}
    \frac{\partial \mathcal{L}_{\mathrm{aux}}}{\partial z_{kj}(t)}
    = \frac{\alpha}{T}\, s_{kj}(t)\Big[\big(1 - e_k(t)\big) - \sum_{i} s_{ij}(t)\big(1 - e_i(t)\big)\Big].
\end{equation}
Since $\sum_i s_{ij}(t) = 1$, the constant cancels and only the error difference remains,
\begin{equation}
\label{equ:AuxGrad}
    \frac{\partial \mathcal{L}_{\mathrm{aux}}}{\partial z_{kj}(t)} = \frac{\alpha}{T}\, s_{kj}(t)\,\big(\bar{e}_j(t) - e_k(t)\big), \qquad \bar{e}_j(t) = \sum_{i} s_{ij}(t)\, e_i(t),
\end{equation}
where $\bar{e}_j(t)$ is the softmax-weighted mean error seen by token $j$.
Treating each logit as a free parameter (logits depend on shared router weights), a gradient step with learning rate lr moves against this gradient,
\begin{equation}
\label{equ:AuxStep}
    \Delta z_{kj}(t) = \mathrm{lr}\,\frac{\alpha}{T}\, s_{kj}(t)\,\big(e_k(t) - \bar{e}_j(t)\big),
\end{equation}
which raises the logit of every expert whose error exceeds $\bar{e}_j(t)$ by an amount linear in that difference and depends on no error before step $t$.
This is the proportional term of \Cref{equ:DiscretePID}, with the token-dependent gain $\mathrm{lr}\,\alpha s_{kj}(t)/T$ in place of the gain $K_P$ and the relative error $e_k(t) - \bar{e}_j(t)$ in place of $e_k(t)$.
The correction does not enter the router as the input $u_i(t)$ of \Cref{equ:Gate} but is added to $g_{\mathrm{LM}}$ (i.e., the language-modeling gradient on the logits) and accumulated with it into the router weights.
The correction itself has no memory, but the accumulation does. 
The update rule alone therefore cannot tell a proportional mechanism from an integral one, which the steady state can.
At a stationary point of the logits under the total objective the two gradients cancel, $g_{\mathrm{LM}} = -\partial \mathcal{L}_{\mathrm{aux}}/\partial z$.
A persistent component of $g_{\mathrm{LM}}$ that favours imbalance therefore acts as a disturbance on the loop, and it is offset only by a nonzero steady-state error that scales as $1/\alpha$ but does not vanish. 
A steady-state error that shrinks with the gain but is never removed is the defining property of proportional control.
We therefore read Aux-loss as a proportional mechanism rather than as an exact instance of \Cref{equ:DiscretePID}.

\paragraph{Auxiliary-Loss-Free Load Balancing \citep{wang2024auxiliary} (ALF-LB) is a \textit{relay-Integral} case.}
ALF-LB measures the error in tokens, $e^{\mathrm{ALF}}_i(t) = r - L_i(t) = r\,e_i(t)$, updates a per-expert bias by
\begin{equation}
\label{equ:ALFLB}
    u_i(t) = u_i(t-1) + \eta\, \operatorname{sign}\big(e^{\mathrm{ALF}}_i(t)\big),
\end{equation}
and injects it exactly as in \Cref{equ:Gate}.
Unrolling \Cref{equ:ALFLB} gives $u_i(t) = \eta \sum_{\tau=0}^{t} \operatorname{sign}\big(e_i(\tau)\big)$ since $\operatorname{sign}(e^{\mathrm{ALF}}_i) = \operatorname{sign}(e_i)$. This is \Cref{equ:DiscretePID} with $K_P = K_D = 0$ and $K_I = \eta$ applied to the sign of the error.
The relay (i.e., $\operatorname{sign}$ in \Cref{equ:ALFLB}) discards the magnitude of the error, and the bias $u_i(t)$ therefore moves by a fixed $\eta$ at every step.
The bias stops drifting only when the error is positive as often as negative, regardless of its magnitude.
Standard integral action removes the steady-state error, and the relay weakens this guarantee to the sign of the error.
\citet{wang2024auxiliary} also examined the linear rule $u_i(t) = u_i(t-1) + \eta\, e^{\mathrm{ALF}}_i(t)$ with a sigmoid gate at one model scale on a fixed data mixture.
They reported better balance at slightly higher perplexity than the relay. 
For the softmax gate of \Cref{equ:Gate}, they made an empirical choice and used this linear rule rather than the relay.
They attributed this choice to the softmax normalization, which makes the score gap between two experts sensitive to the scores of all other experts.

\subsection{What each term contributes}
\label{subsec:terms}

The load of an expert depends on which tokens are in the batch. 
A router with fixed parameters therefore produces a different load at every step.
The resulting error has a slow part and a fast part.
The slow part, an expert staying above or below its share for many consecutive steps, is the persistent error the balancer wants to remove.
The fast part, the step-to-step fluctuation that comes from each step seeing a different batch, is not imbalance but measurement noise.
We examine how each term of \Cref{equ:DiscretePID} responds to the two parts.

\paragraph{Persistent error.}
Consider the extreme case of a dead expert, one that receives no tokens and so has $e_i = 1$ at every step.
The integral term turns this into a ramp that grows by $K_I$ per step, which eventually exceeds any gap in the biased softmax scores and brings the expert back into selection.
The proportional term contributes a constant offset $K_P$ that does not grow, and the expert comes back only if that offset already closes the gap.
It also cannot remove any imbalance completely, since its bias is proportional to the current error.
If the error ever reached zero the bias would be zero as well and the original imbalance would return.
This is the same steady-state error that \Cref{subsec:auxandalflb} identified in Aux-loss.
The derivative term contributes nothing because a constant error has no change.
The same holds for any persistent error $e_i = c$, which only scales the ramp to $K_I c$ per step and the offset to $K_P c$ while leaving the derivative term at zero.

\paragraph{Step-to-step fluctuation.}
Reacting to the fast part is wasted effort, since the bias acts one step late and the fluctuation it responds to is gone by then.
Consider the extreme case of the fastest possible fluctuation, an error that alternates in sign every step, $e_i(t) = (-1)^t$.
By \Cref{equ:DiscretePID} the integral term makes the bias alternate between $0$ and $K_I$ and never changes its sign.
It sums all past errors and each error cancels the one before it, which is how it averages out the fluctuation.
The proportional and derivative terms copy the current error or its change, so their bias flips sign exactly as the error does.
It moves between $-K_P$ and $+K_P$ for the former and between $-2K_D$ and $+2K_D$ for the latter.
Any fluctuation that averages to zero over a few steps behaves the same way.
Its sum cancels within those steps, so only the proportional and derivative terms carry it into the bias.

\paragraph{Consequence.}
An ideal balancer removes the persistent error completely and ignores the step-to-step fluctuation.
The integral term is the only one with accumulated memory and the only one that achieves both.
It adds $K_I e_i(t)$ to the bias at each step and spreads a correction over many steps.
Over those steps the fast part of the error cancels while the slow part accumulates.
Adding either other term brings nothing on the persistent part but adds its swing on the fast part.
We therefore adopt PID (I) with $K_I$ as its only hyperparameter.
\Cref{app:pd} provides more discussion and experiments on the three terms in isolation and in combination.

\section{Experiment}
\label{sec:experiment}
All experiments use the OLMoE backbone \citep{muennighoff2025olmoe} with dropless $\operatorname{Top}\text{-}K$ MoE layers, and \Cref{app:setup} gives the full training and evaluation details.

The PID (I) is evaluated in two regimes and compared with Vanilla (no balance control), Aux-loss and ALF-LB.
In \emph{Pretraining}, the router is trained from scratch with a controlled distribution shift during training, providing an end-to-end view of the router before and after the shift.
In \emph{Continual Pretraining}, a pretrained MoE model is adapted to a narrower target distribution, which tests the PID-Balancer as a drop-in component for existing models.

\paragraph{Metrics.}
Load balancing is only useful if it does not cost language modeling quality, so we report validation perplexity (PPL) as the performance metric that must be preserved.
We adopt MaxVio from \citet{wang2024auxiliary} as the primary balancing metric,
\begin{equation}
\label{equ:MaxVio}
    \mathrm{MaxVio}(t) = \frac{\max_i L_i(t) - r}{r},
\end{equation}
which measures the relative overload of the busiest expert against the perfect balance target $r$ defined in \Cref{equ:Error}.
Since MaxVio only tracks the most overloaded expert, a router can look balanced while starving the under-loaded tail.
We therefore complement it with $R_{\alpha}$
\begin{equation}
\label{equ:Ralpha}
    \mathrm{R}_\alpha(t) = \frac{1}{E} \bigl| \bigl\{ i : L_i(t) < \alpha r \bigr\} \bigr| ,
\end{equation}
which evaluates the load balance from the opposite side by measuring the fraction of experts whose load falls below $\alpha r$.
Setting $\alpha = 0.01$ gives the dead expert ratio $\text{DER}(t) = \mathrm{R}_{0.01}(t)$ and setting $\alpha = 0.25$ gives the under-trained ratio $\text{UTR}(t) = \mathrm{R}_{0.25}(t)$.
All three balancing metrics, MaxVio, DER, and UTR, are computed per MoE layer and reported as the average across layers, and lower is better for each.
We omit the optimization step index $t$ hereafter and report each metric as a scalar averaged over a fixed window. 
PreMV averages MaxVio over the $2{,}000$ steps before the distribution shift, and PostMV averages it over the final $2{,}000$ steps of training.
DER and UTR are averaged over the final $2{,}000$ steps as well.

\subsection{Pretraining}
\label{subsec:pretraining}

We construct a domain-extracted subset of SlimPajama-627B \citep{cerebras2023slimpajama}, keeping 7 predefined source domains (ArXiv, Books, C4, CommonCrawl, GitHub, StackExchange, and Wikipedia) as separate, non-mixed corpora.
The MoE model follows the OLMoE architecture at three scales that vary the number of experts $E \in \{16, 32, 64\}$ while keeping the total and activated parameter counts fixed. 
The routing sparsity is held constant with one eighth of the experts activated per token, so the three scales use $\operatorname{Top}\text{-}2$, $\operatorname{Top}\text{-}4$ and $\operatorname{Top}\text{-}8$ routing respectively.

Each run trains for $33{,}000$ steps from scratch with a single instantaneous domain switch at step $16{,}500$ as the distribution shift. 
We consider three pre-shift mixtures that all switch to GitHub alone.
In \emph{1 $\rightarrow$ 1} the pre-shift data is Books alone. 
In \emph{6 $\rightarrow$ 1} it uniformly weights all domains except GitHub.
In \emph{7 $\rightarrow$ 1} it uniformly weights all seven domains including GitHub.
Results below are averaged over 6 runs with different random seeds, with error bars denoting the standard deviation across seeds.
The vanilla baseline has no load balancing and uses a single seed as a reference only.

\begin{figure}[t]
\begin{center}
\includegraphics[width=0.85\textwidth]{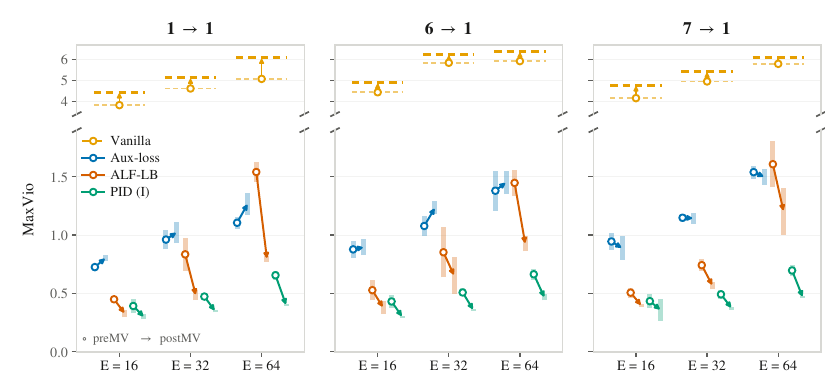}
\end{center}
\caption{Pre- and post-shift MaxVio for the three pre-shift scenarios of \Cref{subsec:pretraining}. The circle marks the transition from the pre-shift steady state (preMV) to the post-shift steady state (postMV).} 
\label{fig:maxvio_phase2}
\end{figure}

\begin{figure}[t]
\begin{center}
\includegraphics[width=0.8\textwidth, trim=10pt 5pt 2pt 7pt, clip]{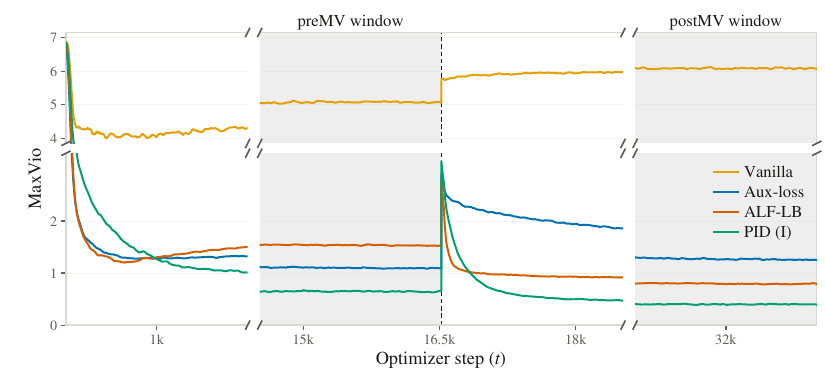}
\end{center}
\caption{MaxVio over training (E = 64, \textit{1 $\rightarrow$ 1} scenario, 6-seed mean, 51-step smoothing). Shading marks the 2,000-step windows averaged for PreMV and PostMV.}
\label{fig:maxvio_traj}
\end{figure}

\paragraph{Results.}
\Cref{tab:pretraining_results} and \Cref{fig:maxvio_phase2} summarize the metrics in the \emph{pretraining} regime, and \Cref{fig:maxvio_traj} shows an example MaxVio trajectory illustrating how the shift evolves over training.
Across all scenarios and scales, the three balancers adapt to the distribution shift far more effectively than the uncontrolled Vanilla baseline, settling at a markedly lower post-shift MaxVio and largely eliminating the dead and under-trained experts that Vanilla leaves behind.
Aux-loss balances considerably better than Vanilla but less effectively than the two bias-based methods, with a few under-trained experts remaining in some runs.
ALF-LB improves balance substantially after the shift, but both its pre- and post-shift MaxVio fall further behind PID as the number of experts grows.
The proposed PID (I) achieves the lowest MaxVio both before and after the shift in every setting.
Perplexity remains nearly identical across balancers in every setting, indicating that the improved balance does not compromise language-modeling quality.

\begin{figure}[t]
\begin{center}
\includegraphics[width=0.99\textwidth]{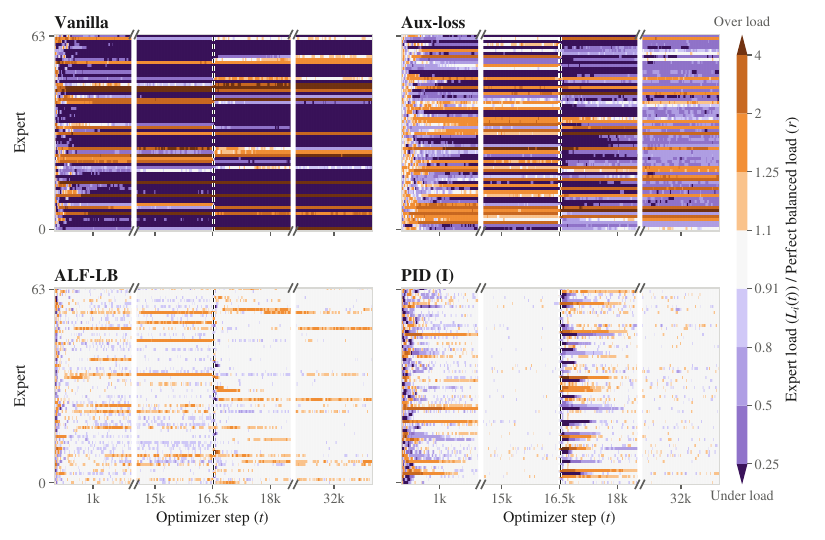}
\end{center}
\caption{Per-expert load comparison.
Color encodes the load of an expert divided by the perfectly balanced load.
Rows are the $64$ experts of the last MoE layer, columns are optimizer steps, and the dashed line marks the mixture shift from Books to GitHub at step $16{,}500$.
The $x$-axis is broken to show the initial transient, the pre-shift steady state, the post-shift transient, and the end of training.}
\label{fig:expertbalance}
\end{figure}

\Cref{fig:expertbalance} visualizes a per-expert load example at $E = 64$ on the \emph{1 $\rightarrow$ 1} scenario.
Vanilla and Aux-loss largely keep a set of over- and under-loaded experts on both sides of the distribution shift, and the shift mostly changes which experts these are.
Both bias-based methods keep the load much closer to the balanced share during the steady phases, although ALF-LB retains a few mildly overloaded experts while PID (I) stays nearly uniform.
PID (I) shows a short burst of imbalance right after the shift because its biases fit the pre-shift distribution most closely and therefore suffer the largest mismatch. 
The burst decays quickly and PID (I) ends training with the most even load of all methods.

\begin{table}[t]
\caption{Post-shift steady-state comparison across expert count $E \in \{16, 32, 64\}$ and the three pre-shift mixture scenarios of \Cref{subsec:pretraining}. Arrows show the better direction ($\uparrow$ higher, $\downarrow$ lower).}
\label{tab:pretraining_results}
\begin{center}
\resizebox{\textwidth}{!}{%
\begin{tabular}{cl ccc ccc ccc}
\toprule
\multicolumn{2}{c}{} & \multicolumn{3}{c}{\textbf{1 $\rightarrow$ 1}} & \multicolumn{3}{c}{\textbf{6 $\rightarrow$ 1}} & \multicolumn{3}{c}{\textbf{7 $\rightarrow$ 1}} \\
\cmidrule(lr){1-2} \cmidrule(lr){3-5} \cmidrule(lr){6-8} \cmidrule(lr){9-11}
\textbf{E} & \textbf{Method} & postMV $\downarrow$ & PPL $\downarrow$ & DER $\downarrow$ / UTR $\downarrow$ & postMV $\downarrow$ & PPL $\downarrow$ & DER $\downarrow$ / UTR $\downarrow$ & postMV $\downarrow$ & PPL $\downarrow$ & DER $\downarrow$ / UTR $\downarrow$ \\
\midrule
\multirow{4}{*}{16}
 & Vanilla & 4.44 & 3.29 & 7.42\% / 38.67\% & 4.89 & 3.22 & 5.08\% / 41.41\% & 4.77 & 3.21 & 5.86\% / 37.89\% \\
 & Aux-loss & $0.80 \pm 0.03$ & $3.31 \pm 0.01$ & 0.00\% / 0.23\% & $0.90 \pm 0.07$ & $3.23 \pm 0.02$ & 0.00\% / 0.00\% & $0.89 \pm 0.10$ & $3.20 \pm 0.02$ & 0.00\% / 0.36\% \\
 & ALF-LB   & $0.33 \pm 0.03$ & $3.32 \pm 0.01$ & 0.00\% / 0.00\% & $0.38 \pm 0.06$ & $3.25 \pm 0.01$ & 0.00\% / 0.00\% & $0.40 \pm 0.01$ & $3.24 \pm 0.03$ & 0.00\% / 0.00\% \\
 & PID (I)  & $\boldsymbol{0.30 \pm 0.02}$ & $3.32 \pm 0.01$ & 0.00\% / 0.00\% & $\boldsymbol{0.30 \pm 0.01}$ & $3.25 \pm 0.02$ & 0.00\% / 0.00\% & $\boldsymbol{0.36 \pm 0.09}$ & $3.23 \pm 0.02$ & 0.00\% / 0.00\% \\
\midrule
\multirow{4}{*}{32}
 & Vanilla & 5.13 & 3.31 & 12.30\% / 44.34\% & 6.23 & 3.21 & 9.96\% / 51.76\% & 5.42 & 3.21 & 9.18\% / 42.58\% \\
 & Aux-loss & $1.02 \pm 0.09$ & $3.29 \pm 0.02$ & 0.00\% / 0.26\% & $1.23 \pm 0.06$ & $3.21 \pm 0.01$ & 0.00\% / 0.00\% & $1.15 \pm 0.05$ & $3.21 \pm 0.01$ & 0.00\% / 1.26\% \\
 & ALF-LB   & $0.48 \pm 0.04$ & $3.32 \pm 0.02$ & 0.00\% / 0.00\% & $0.65 \pm 0.16$ & $3.25 \pm 0.01$ & 0.00\% / 0.00\% & $0.57 \pm 0.03$ & $3.23 \pm 0.01$ & 0.00\% / 0.00\% \\
 & PID (I)  & $\boldsymbol{0.35 \pm 0.01}$ & $3.31 \pm 0.02$ & 0.00\% / 0.00\% & $\boldsymbol{0.36 \pm 0.01}$ & $3.23 \pm 0.01$ & 0.00\% / 0.00\% & $\boldsymbol{0.37 \pm 0.01}$ & $3.21 \pm 0.03$ & 0.00\% / 0.00\% \\
\midrule
\multirow{4}{*}{64}
 & Vanilla & 6.08 & 3.30 & 16.31\% / 45.41\% & 6.37 & 3.23 & 19.92\% / 48.63\% & 6.10 & 3.23 & 23.54\% / 56.05\% \\
 & Aux-loss & $1.27 \pm 0.10$ & $3.29 \pm 0.02$ & 0.06\% / 0.21\% & $1.45 \pm 0.10$ & $3.21 \pm 0.01$ & 0.00\% / 0.73\% & $1.50 \pm 0.07$ & $3.21 \pm 0.03$ & 0.00\% / 1.09\% \\
 & ALF-LB   & $0.80 \pm 0.03$ & $3.31 \pm 0.02$ & 0.00\% / 0.00\% & $0.92 \pm 0.06$ & $3.24 \pm 0.01$ & 0.00\% / 0.00\% & $1.20 \pm 0.20$ & $3.22 \pm 0.02$ & 0.00\% / 0.00\% \\
 & PID (I)  & $\boldsymbol{0.40 \pm 0.01}$ & $3.30 \pm 0.02$ & 0.00\% / 0.00\% & $\boldsymbol{0.47 \pm 0.03}$ & $3.24 \pm 0.01$ & 0.00\% / 0.00\% & $\boldsymbol{0.47 \pm 0.01}$ & $3.21 \pm 0.02$ & 0.00\% / 0.00\% \\
\bottomrule
\end{tabular}}
\end{center}
\end{table}

\subsection{Continual Pretraining}
\label{subsec:continualpretraining}
Continual pretraining (CPT) starts from the released OLMoE checkpoint \texttt{OLMoE-1B-7B-0125}.
This checkpoint was first pretrained on \texttt{OLMoE-mix-0924} and then annealed on the Dolmino mixture \citep{walsh2025}, which we refer to as the \emph{annealing corpus}.
We continue training this checkpoint on \texttt{codeparrot} \citep{tunstall2022natural}, which we refer to as the \emph{CPT corpus}.

Since code already appears in OLMoE's pretraining mixture, this step changes the mixture ratio rather than moving to an unseen domain.
It is therefore structurally analogous to the \emph{7 $\rightarrow$ 1} scenario in \Cref{subsec:pretraining}.
To study how each balancer behaves as a drop-in replacement on an already pretrained model, we attach it to the checkpoint and consider two start conditions.
In the \emph{cold-start} setting, the model is trained directly on the CPT corpus for $16{,}500$ steps.
In the \emph{warm-start} setting, the model is first trained on the Annealing corpus for $16{,}500$ steps so that the balancer reaches a steady state.
Training then switches to the CPT corpus for another $16{,}500$ steps.

In addition to the balancing metrics, we measure perplexity on two held-out sets.
The first is drawn from the CPT corpus and measures adaptation to the new domain, while the second is drawn from the Annealing corpus and measures forgetting of the prior distribution.
Both perplexities are evaluated every $250$ steps, and we report the mean over the last eight evaluation points of training, which corresponds to the $2{,}000$-step averaging window defined in \Cref{sec:experiment}.
We also evaluate five downstream benchmarks (MMLU, HellaSwag, ARC-Challenge, HumanEval and MBPP) on the base checkpoint and on each method at the end of continual pretraining.

\paragraph{Results.} 
\Cref{tab:cpt} reports load balancing metrics and task performance for the four methods.
In both settings, Vanilla shows a strongly imbalanced load, settling on the CPT corpus at a postMV about six times that of the bias-based methods ALF-LB and PID (I).
Aux-loss only partially reduces this imbalance, and its postMV remains more than four times that of the bias-based methods.
Raising the Aux-loss coefficient does reduce postMV, but every gain in balance is paid for in perplexity and downstream accuracy (\Cref{app:p3auxloss}).
The proposed PID (I) attains the lowest postMV in both settings.
In the warm-start setting, the same ranking already holds before the switch.
The low preMV of Vanilla is expected, since the checkpoint was trained on the Annealing corpus with an auxiliary loss.
Vanilla therefore starts from an already well-balanced state on this corpus, and the balance is lost once the corpus changes.

Adaptation to the new domain is essentially insensitive to the choice of balancer.
Within each setting, the four methods reach nearly the same held-out perplexity on the CPT corpus (PPL-CPT), and all of them improve on the base checkpoint.
Forgetting behaves differently.
The two bias-based methods balance far better, but they also forget more within the same training budget.
In both settings, their held-out perplexity on the Annealing corpus (PPL-Annealing) is higher than that of Vanilla and Aux-loss, and their downstream scores are slightly lower or similar.
This trade-off is not surprising.
A stronger balancer pushes tokens onto experts that the router would not have chosen.
We believe experts specializing in the prior domain are more likely to receive these force-reassigned tokens on the new corpus, which could overwrite what they had learned.

The results from the two start settings are not directly comparable, since the warm-start runs have seen $16{,}500$ additional steps on the Annealing corpus.
This accounts for their higher scores on MMLU, HellaSwag and ARC-C, and the same gap appears even for Vanilla baseline.
Even though the router in the cold-start checkpoint has never included these balancers before, it reaches a postMV close to that of the warm-start setting.
Therefore, both PID (I) and ALF-LB can serve as drop-in replacements on an existing pretrained model.

\begin{table}[t]
\caption{Continual pretraining results of four routing methods under two initializations.}
\label{tab:cpt}
\begin{center}
\resizebox{\textwidth}{!}{%
\begin{tabular}{llrrrrrrrrrrr}
\toprule
Method & Setting & preMV$\downarrow$ & postMV$\downarrow$ & DER$\downarrow$ / UTR$\downarrow$ & PPL-CPT$\downarrow$ & PPL-Annealing$\downarrow$ & MMLU$\uparrow$ & HellaSwag$\uparrow$ & ARC-C$\uparrow$ & HumanEval$\uparrow$ & MBPP$\uparrow$ \\
\midrule
Base & - & - & - & - & 2.40 & 5.46 & 0.55 & 0.79 & 0.57 & 0.11 & 0.23 \\
\midrule
Vanilla & cold & - & 5.83 & 1.07\% / 29.10\% & 2.19 & 7.23 & 0.50 & 0.74 & 0.50 & 0.12 & 0.18 \\
Aux-loss & cold & - & 4.43 & 0.00\% / 1.56\% & 2.19 & 7.21 & 0.49 & 0.74 & 0.50 & 0.12 & 0.18 \\
ALF-LB & cold & - & 1.01 & 0.00\% / 0.00\% & 2.19 & 7.36 & 0.46 & 0.71 & 0.48 & 0.10 & 0.14 \\
PID (I) & cold & - & \textbf{0.90} & 0.00\% / 0.00\% & 2.19 & 7.39 & 0.46 & 0.72 & 0.50 & 0.12 & 0.14 \\
\midrule
Vanilla & warm & 1.14 & 5.76 & 0.59\% / 27.73\% & 2.24 & 6.07 & 0.51 & 0.77 & 0.55 & 0.12 & 0.18 \\
Aux-loss & warm & 1.09 & 4.51 & 0.00\% / 6.93\% & 2.24 & 6.07 & 0.51 & 0.77 & 0.56 & 0.13 & 0.19 \\
ALF-LB & warm & 0.72 & 1.03 & 0.00\% / 0.00\% & 2.24 & 6.28 & 0.48 & 0.75 & 0.54 & 0.10 & 0.13 \\
PID (I) & warm & 0.66 & \textbf{0.87} & 0.00\% / 0.00\% & 2.23 & 6.28 & 0.48 & 0.75 & 0.55 & 0.11 & 0.13 \\
\bottomrule
\end{tabular}%
}
\end{center}
\end{table}

\section{Discussion}
\label{sec:discussion}

\paragraph{Better balance, better efficiency.}
\citet{therien2025continual} propose Maximum Routing Imbalance (MRI), the largest proportion of tokens routed to a single expert in a given MoE layer, as a proxy for a layer's worst-case latency under expert parallelism.
Since $\text{MRI} = (\text{MaxVio}+1)\cdot K/E$ (\Cref{app:maxvio_mri}), every balance result in this paper also converts to an MRI estimate that approximates this latency proxy.
However, the latency reading remains conditional on deployment, so we take load balance as our primary metric.

\paragraph{Trade-off between balance and specialization.}
Routing and expert knowledge shape each other during training \citep{guo2025advancing}, so tokens of the same kind gather on a few experts.
In OLMoE, some experts receive almost all tokens from certain domains or vocabulary items, while others stay far below the uniform-routing baseline \citep{muennighoff2025olmoe}.
Since different kinds of knowledge are not needed equally often, these experts will not receive equal load.
Equalizing load therefore sends some of these tokens to experts that lack the matching knowledge \citep{guo2025advancing}.
The OLMoE authors note that their auxiliary loss forces near-equal expert use and may prevent experts from specializing in certain domains \citep{muennighoff2025olmoe}.

Unlike Aux-loss, which balances load through the gradient on the gate weights, bias-based balancers such as ours act only on expert selection.
Our specialization probe (\Cref{app:specialization}) shows that this difference matters.
Aux-loss reduces the dependence of routing on the input domain, yet its load remains less balanced than that of ALF-LB and PID (I).
This agrees with \citet{guo2025advancing}, who attribute the loss of specialization under auxiliary losses to tokens being routed to experts that do not match them.
In ALF-LB and PID (I), the bias produces the same mismatch only at selection time by overriding the domain preference of the gate.
However, the gate weights themselves retain most of their domain preference.

\paragraph{Additional results.}
\Cref{app:knob} varies the balance strength coefficient of each method and presents the resulting trade-off curve between balance and perplexity.
For pretraining under the $1\!\to\!1$ scenario, \Cref{app:ki} sweeps the gain $K_I$ and gives an empirical rule for transferring it across expert counts rather than re-tuning it.
Routing sparsity is ablated in \Cref{app:topk}, which shows the same pattern as \Cref{subsec:pretraining}.
Continual pretraining is extended to longer budgets in \Cref{app:cptmore} and reaches the same conclusions as \Cref{subsec:continualpretraining}.
The layer-wise load balance is analyzed in \Cref{app:layerwise}.
Limitations and future work are discussed in \Cref{app:limitation}.

\section{Conclusion}
\label{sec:conclusion}

We have shown that MoE load balancing can be viewed as a feedback control problem, and that two widely used balancers belong to a single PID family.
The auxiliary loss acts as a proportional term through the gate weights, while the loss-free bias update of ALF-LB acts as a relay integrator on the routing decision.
The control view provides a unified setup in which all three terms can be analyzed, including the derivative term that no existing balancer uses.
Under this setup, the step-to-step load error is dominated by micro-batch noise.
The proportional and derivative terms pass this noise into the bias, while the integral term averages it out.
This points to a linear integrator on the normalized load error, which achieves the lowest steady-state imbalance among all compared methods in pretraining and continual pretraining, with comparable perplexity and downstream accuracy.

\subsubsection*{Acknowledgments}
This work is supported by the Wallenberg AI, Autonomous Systems, and Software Program (WASP), funded by the Knut and Alice Wallenberg Foundation.
Computational resources were provided on the Berzelius system funded by the Knut and Alice Wallenberg Foundation and operated by NAISS.

\bibliography{references}
\bibliographystyle{plainnat}

\clearpage
\newpage

\appendix

\section{Pseudocode of PID-Balancer}
\label{app:pseudocode}

Each of the $L$ MoE layers runs an independent instance of \Cref{alg:pidbalancer}.
All layers share the same gains $(K_P, K_I, K_D)$, but each layer keeps its own per-expert state.
Consistent with the main text, the layer index is dropped below for clarity.

\begin{algorithm}[h]
\caption{PID load-balancing update, executed once per optimizer step}
\label{alg:pidbalancer}
\begin{algorithmic}[1]
\Require gains $K_P, K_I, K_D$, setpoint $r = TK/E$
\State \textbf{Persistent state:} integral $I_i$, previous error $e_i^{\text{prev}}$, bias $u_i$
\State Initialize $I_i, e_i^{\text{prev}}, u_i \gets 0$ for all experts $i$
\For{each optimizer step $t = 1, 2, \dots$}
    \State $L_i \gets$ tokens routed to expert $i$ during step $t$
    \State $e_i \gets (r - L_i) / r$ \Comment{\Cref{equ:Error}}
    \State $I_i \gets I_i + K_I\, e_i$
    \State $D_i \gets K_D\, (e_i - e_i^{\text{prev}})$
    \State $u_i \gets K_P\, e_i + I_i + D_i$ \Comment{absolute bias, \Cref{equ:DiscretePID}, overwrites the previous $u_i$}
    \State $e_i^{\text{prev}} \gets e_i$
    \For{every token $j$ routed by this layer at step $t+1$}
        \State $\mathcal{T}_j \gets \operatorname{Top}\text{-}K_i\big(s_{ij} + u_i\big)$ \Comment{\Cref{equ:Gate}}
    \EndFor
    \State $L_i \gets 0$ for all $i$ \Comment{reset the load accumulator}
\EndFor
\end{algorithmic}
\end{algorithm}

\paragraph{Zeroing individual gains.} 
Setting any subset of $K_P, K_I, K_D$ to zero switches off the corresponding term without altering \Cref{alg:pidbalancer} itself. 
Our main method PID (I) is a pure linear integral balancer, and we obtain it this way by setting $K_P = K_D = 0$.

\paragraph{Initialization of the derivative term.} 
$D_i$ needs two consecutive error observations and is therefore undefined at the first optimizer step. 
For derivative term only, we handle this at initialization by setting $e_i^{\text{prev}}$ to $e_i$ at $t=1$, which is equivalent to $D_i(1) := 0$.
This prevents a false derivative response at the first step, since such a response would come from the zero initial condition and not from an observed rate of change.

\section{Additional Experimental Setup}
\label{app:setup}

\paragraph{Computational Resources and Environment:} All experiments were conducted using Python 3.11.15 and PyTorch 2.7.1 on NVIDIA four A100 GPUs.

\paragraph{Balancer coefficient.}
We compare PID-Balancer against three baselines. 
\textbf{Vanilla} is $\operatorname{Top}\text{-}K$ routing with no load-balancing mechanism and serves as the baseline reference. 
\textbf{Auxiliary loss (Aux-loss)} is the Switch-Transformer-style soft load-balancing loss \citep{fedus2022switch}, with coefficient $\alpha =$1e-2.
\textbf{Auxiliary-Loss-Free Load Balancing (ALF-LB)} is the bias-based rule of \citet{wang2024auxiliary}, with coefficient $\eta =$ 1e-3.
We sweep both coefficients in \Cref{app:knob}. 

\paragraph{Model and data configurations.}
\Cref{apptab:model} lists the architecture details for each model scale. 
\Cref{apptab:data} lists the corpora and training setup used for pretraining and continual pretraining.

\begin{table}[t]
\caption{Model architecture and MoE configuration across scales. The pretraining experiment uses OLMoE-16/32/64, and the continual pretraining experiment uses OLMoE-1B-7B.}
\label{apptab:model}
\begin{center}
\renewcommand{\arraystretch}{1.3}
\resizebox{\textwidth}{!}{%
\begin{tabular}{p{3.6cm}cccc}
\hline
 & \textbf{OLMoE-16} & \textbf{OLMoE-32} & \textbf{OLMoE-64} & \textbf{OLMoE-1B-7B} \\
\hline
\multicolumn{5}{l}{\textit{Backbone}} \\
\hline
Hidden size & 1280 & 1280 & 1280 & 2048 \\
Layers (all MoE) & 16 & 16 & 16 & 16 \\
Attention heads & 20 & 20 & 20 & 16 \\
Head dim & 64 & 64 & 64 & 128 \\
Normalisation & RMSNorm, $\varepsilon=$ 1e-5 & RMSNorm, $\varepsilon=$ 1e-5 & RMSNorm, $\varepsilon=$ 1e-5 & RMSNorm, $\varepsilon=$ 1e-5 \\
QK-norm & off & off & off & on \\
Vocabulary & 50304 & 50304 & 50304 & 50304 \\
\hline
\multicolumn{5}{l}{\textit{Mixture of experts}} \\
\hline
Experts \texttt{E} & 16 & 32 & 64 & 64 \\
$\operatorname{Top}\text{-}K$ & 2 & 4 & 8 & 8 \\
Sparsity $\operatorname{Top}\text{-}K/E$ & 1/8 & 1/8 & 1/8 & 1/8 \\
Expert hidden size & 2048 & 1024 & 512 & 1024 \\
Dropless routing & yes & yes & yes & yes \\
Gating & softmax & softmax & softmax & softmax \\
Total parameters & 2.25B & 2.25B & 2.25B & 6.92B \\
Active parameters & 0.49B & 0.49B & 0.49B & 1.28B \\
\hline
\end{tabular}
}
\end{center}
\end{table}

\begin{table}[t]
\caption{Pretraining and continual pretraining setup}
\label{apptab:data}
\begin{center}
\renewcommand{\arraystretch}{1.3}
\resizebox{\textwidth}{!}{%
\begin{tabular}{p{3cm}p{7.5cm}p{7.5cm}}
\hline
 & \textbf{Pretraining} & \textbf{Continual pretraining} \\
\hline
\multicolumn{3}{l}{\textit{\textbf{Corpus}}} \\
\hline
Corpus & SlimPajama-627B & Anneal: \texttt{Dolmino} \newline CPT: \texttt{codeparrot-clean} \\
\hline
Domains & 7 domains (arxiv, Books, c4, commoncrawl, GitHub, stackexchange, wikipedia) & Anneal: 6 domains (dclm, flan, math, pes2o, stackexchange, wiki). \newline CPT: 1 domain (code) \\
\hline
Tokenizer & \texttt{gpt-neox-olmo-dolma-v1\_5} & OLMoE-native tokenizer \\
\hline
Vocabulary / EOS / BOS & 50,280 (embedding padded to 50,304), EOS 50279, no BOS & 50,280 (embedding padded to 50,304), EOS 50279, no BOS \\
\hline
\multicolumn{3}{l}{\textit{\textbf{Training consumption}}} \\
\hline
Mixture mechanism & All non-target domains mixed uniformly (depends on the scenario), target domain is \texttt{GitHub} & All non-target domains mixed uniformly, target domain is code (CPT) \\
\hline
Sequence length & 2,048 & 4,096 \\
\hline
Global batch & 16 sequences & 8 sequences \\
\hline
Batch tokens & 32,768 & 32,768 \\
\hline
Total steps & 33,000 & 33,000 (warm-start), 16,500 (cold-start) \\
\hline
\multicolumn{3}{l}{\textit{\textbf{Evaluation corpora}}} \\
\hline
Held-out sets & None & Anneal held-out (20M tokens) measures \textit{forgetting}. \newline CPT held-out (20M tokens) measures \textit{adaptation} \\
\hline
Benchmarks & None & \texttt{mmlu}, \texttt{hellaswag}, \texttt{arc\_challenge}, \texttt{humaneval}, \texttt{mbpp} \\
\hline
\multicolumn{3}{l}{\textit{\textbf{Optimizer}}} \\
\hline
Optimizer & AdamW & AdamW \\
\hline
Lr & 4.00e-4 & 6.14e-05 \\
\hline
LR schedule & None & None \\
\hline
Betas / eps & (0.9, 0.999) / 1e-8 & (0.9, 0.999) / 1e-8 \\
\hline
Weight decay & 0.01 & 0.0 \\
\hline
Gradient clipping & Gradient norm clipped to 1.0 & Gradient norm clipped to 1.0 \\
\hline
\multicolumn{3}{l}{\textit{\textbf{Reproducibility}}} \\
\hline
Seed & 42, with additional 5 random seeds generated from it & 42 \\
\hline
\end{tabular}
}
\end{center}
\end{table}

\paragraph{Benchmark evaluation for continual pretraining.}
We evaluate on five benchmarks: MMLU, HellaSwag, ARC-Challenge, HumanEval, and MBPP, using the EleutherAI lm-evaluation-harness (v0.4.12).
MMLU is scored with accuracy, HellaSwag and ARC-Challenge with length-normalized accuracy, and the two code tasks with pass@1. 
Pass@1 is computed from a single greedy generation per problem, taking the mean of the resulting per-problem 0/1 outcomes.
MMLU, HellaSwag, and ARC-Challenge use 5-shot evaluation, passed explicitly, while HumanEval (0-shot) and MBPP (3-shot) use the default task configurations of the harness.
Batch size is 16 for the loglikelihood tasks (MMLU, HellaSwag, ARC-Challenge) and 12 for the generative tasks (HumanEval, MBPP).
All runs use bf16 precision and a fixed seed of 42 for all random number generators, including few-shot sampling.

\section{Additional Experiments}
\subsection{Effect of the Balance Strength Coefficient}
\label{app:knob}

Each balancer evaluated in this study has a single balance strength scalar that controls how hard it pushes against load imbalance.
We sweep that scalar for every method and report balance (MaxVio) and task performance (PPL) jointly.
Presented experiments here use the \emph{1 $\rightarrow$ 1} scenario with $E = 64$ and $\operatorname{Top}\text{-}8$, following the configuration of \Cref{subsec:pretraining}.
For Aux-loss, the scalar is the coefficient $\alpha$ on the auxiliary term.
For ALF-LB, it is the bias update rate $\eta$.
For PID (I), it is the integral gain $K_I$.
All three are swept over the same grid, $\{$1e-5, 1e-4, 1e-3, 1e-2, 1e-1$\}$.
Additionally, setting any of the three scalars to zero recovers the uncontrolled vanilla method.
At $\alpha = 0$ the auxiliary term contributes no gradient and the optimized objective is the task cross-entropy alone.
At $\eta = 0$ or $K_I = 0$ the bias never leaves its zero initialization, so every selection is taken on the raw router logits.

\begin{figure}[t]
    \begin{center}
    \includegraphics[width=0.9\textwidth]{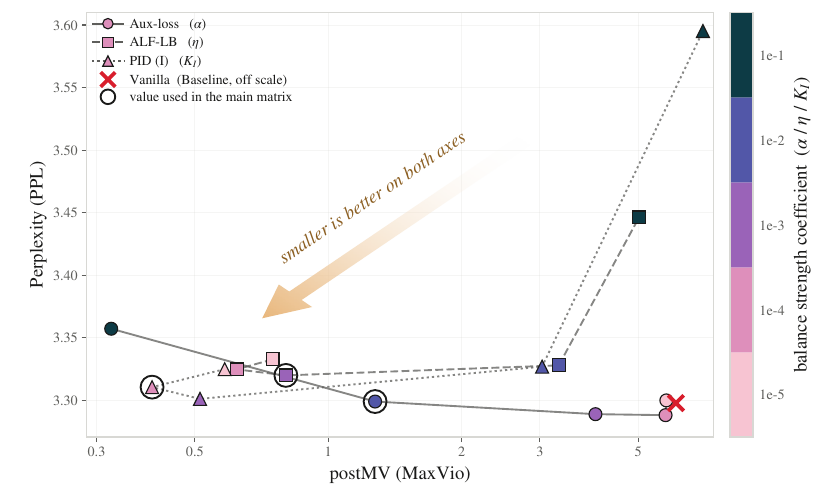}
    \end{center}
    \caption{Load balance and task performance across the swept balance strength coefficient for each method.}
    \label{appfig:knob}
\end{figure}

\Cref{appfig:knob} illustrates the trade-off curve between load balance and task performance for each method.
We therefore adopt $\alpha = $1e-2 for Aux-loss throughout the experiments, balancing task performance against load-balancing.
Since results at $\eta = $1e-3 and $\eta = $1e-4 are indistinguishable in ALF-LB, we adopt $\eta = $1e-3, which is also the value recommended in the source paper \citep{wang2024auxiliary}.

Across all three methods, the balance strength setting that minimizes MaxVio never coincides with the setting that minimizes PPL.
For Aux-loss, postMV decreases monotonically as $\alpha$ grows across the swept range, with the best balance at the largest value tested, $\alpha=$1e-1, while PPL reaches its minimum at $\alpha=$1e-4.
ALF-LB and PID (I) are non-monotone on both axes and share the same shape.
Balance is minimized at an interior gain and perplexity is minimized one grid step higher.
This holds at $\eta=$1e-4 versus $\eta=$1e-3 for ALF-LB, and at $K_I=$1e-4 versus $K_I=$1e-3 for PID (I).
This separation suggests that pushing for uniform load and routing tokens to their loss-preferred experts are competing pressures, not a single objective that can be jointly optimized.

At large gain, the two bias-based methods degrade language modeling performance, but ALF-LB preserves it better.
This gap comes from how each method uses the accumulated imbalance signal.
ALF-LB scales its correction by the sign of the imbalance rather than its magnitude, which keeps the update bounded even when the gain is large.
PID (I) instead multiplies the full integral value by the gain. 
As a result, the correction grows with both the error magnitude and the gain, which drives the relative instability at large $K_I$.
Using the sign therefore trades away information about the imbalance magnitude for a more stable operating range.

Aux-loss and the two bias-based methods fail on balance differently at their worst settings.
DER (Dead Expert Ratio) and UTR (Under-Trained Ratio), introduced in \Cref{sec:experiment}, reach 15.14\% and 44.73\% respectively at $\alpha=$1e-5, the setting where Aux-loss's MaxVio is worst.
The high MaxVio there flags overloaded experts, and DER and UTR flag starved experts. 
The two statistics therefore point at the same failure.

At their own most aggressive setting, the two bias methods fail differently.
PostMV there is comparably high to Aux-loss's worst setting, yet DER and UTR remain at zero there. 
This holds across almost the entire sweep, with a small nonzero UTR appearing only at the smallest tested value.
MaxVio is a per-optimization-step, single-expert statistic. 
With $E=64$ and $\operatorname{Top}\text{-}8$, a MaxVio of 7 means one expert was selected by every token in that step. 
This expert occupies only one of the eight slots per token and leaves the other seven free, not that only 8 experts were selected in total.
For the bias methods it is not the same expert.
Zero DER and UTR indicate that no expert is left idle over the statistical window (2,000 steps in this case), meaning the overloaded expert rotates over time while the window-averaged utilization stays even.

The sweep also reveals this asymmetry in how PPL and postMV degrade across the sweep.
Aux-loss degrades gradually across the full range, with perplexity rising only mildly from best to worst and postMV staying below the Vanilla baseline throughout.
The two bias-based methods instead remain roughly flat for most of the sweep, but their load balance degradation only rises sharply at the largest tested value.

\begin{table}[t]
\caption{DER and UTR across balance strength coefficients for the three balancers. * indicates the value used in the main text.}
\label{tab:der_utr_shared}
\begin{center}
\newcommand{\ctr}[1]{\multicolumn{1}{c}{#1}}
\begin{tabular}{lllrr}
\toprule
Method & Coefficient & Value & DER & UTR \\
\midrule
\multirow{5}{*}{Aux-loss} & \multirow{5}{*}{$\alpha$} & 1e-5 & 15.14\% & 44.73\% \\
 & & 1e-4 & 16.21\% & 43.26\% \\
 & & 1e-3 & 3.61\% & 24.02\% \\
 & & 1e-2* & 0.06\% & 0.21\% \\
 & & 1e-1 & 0.00\% & 0.00\% \\
\midrule
\multirow{5}{*}{ALF-LB} & \multirow{5}{*}{$\eta$} & 1e-5 & 0.00\% & 0.49\% \\
 & & 1e-4 & 0.00\% & 0.00\% \\
 & & 1e-3* & 0.00\% & 0.00\% \\
 & & 1e-2 & 0.00\% & 0.00\% \\
 & & 1e-1 & 0.00\% & 0.00\% \\
\midrule
\multirow{5}{*}{PID (I)} & \multirow{5}{*}{$K_I$} & 1e-5 & 0.10\% & 0.39\% \\
 & & 1e-4* & 0.00\% & 0.00\% \\
 & & 1e-3 & 0.00\% & 0.00\% \\
 & & 1e-2 & 0.00\% & 0.00\% \\
 & & 1e-1 & 0.00\% & 0.00\% \\
\bottomrule
\end{tabular}
\end{center}
\end{table}

\subsection{Choice of the integral gain}
\label{app:ki}

\begin{figure}[t]
    \begin{center}
    \includegraphics[width=0.99\textwidth]{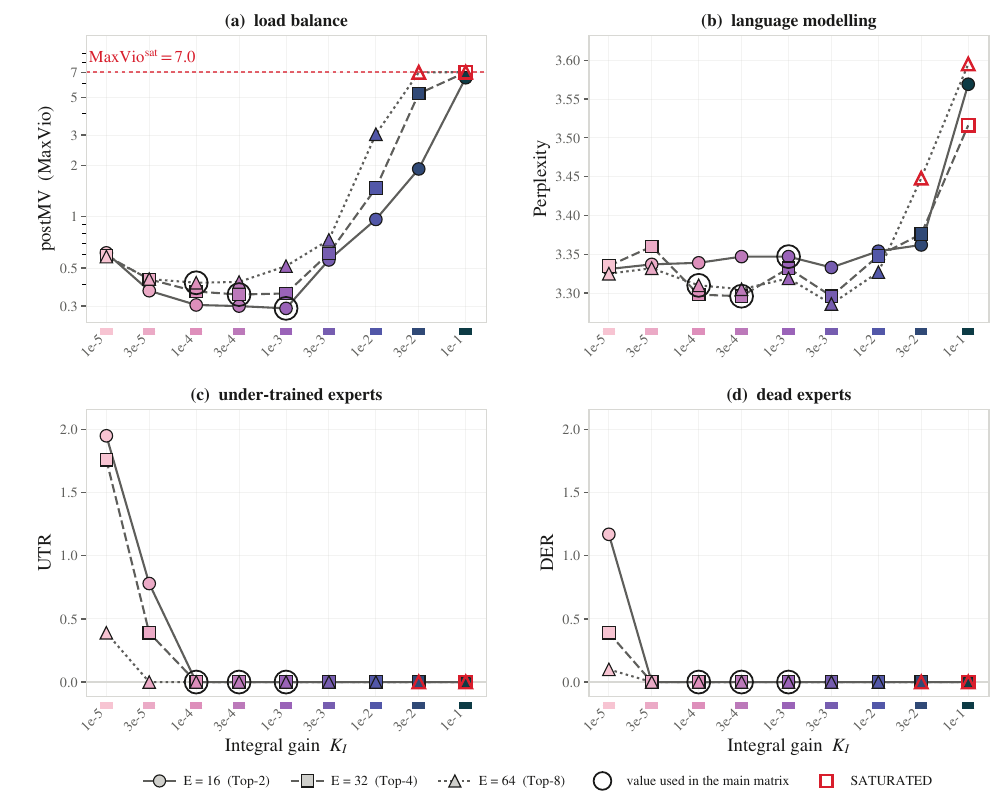}
    \end{center}
    \caption{$K_I$ sweep performance under the \emph{1 $\rightarrow$ 1} scenario at three scales, $E = 16$ with $\operatorname{Top}\text{-}2$, $E = 32$ with $\operatorname{Top}\text{-}4$, and $E = 64$ with $\operatorname{Top}\text{-}8$.
    Lower is better in every panel.
    The saturation ceiling is defined in \Cref{app:topk}.}
    \label{appfig:kisweep}
\end{figure}

The integral gain $K_I$ is the only free hyperparameter of PID (I), and here we report the sweep it was chosen from.
The grid takes nine values as shown in \Cref{appfig:kisweep}, repeated at three scales under the \emph{1 $\rightarrow$ 1} scenario.
The scales are $E = 16$ with $\operatorname{Top}\text{-}2$, $E = 32$ with $\operatorname{Top}\text{-}4$, and $E = 64$ with $\operatorname{Top}\text{-}8$.
All three hold the routing sparsity fixed at $1/8$, so the MaxVio saturation ceiling is $16/2 - 1 = 7.0$, $32/4 - 1 = 7.0$, and $64/8 - 1 = 7.0$ respectively (\Cref{app:topk}).

Load balance is U-shaped in $K_I$ at every scale.
Moving it by two orders of magnitude takes postMV from below $0.30$ to the ceiling.
The two ends fail by different mechanisms.
At the smallest gains the integrator is too slow to prompt experts the router has stopped selecting, and UTR and DER are non-zero at five sweep points, all with $K_I \le $3e-5.
At the largest gains the integrator is driven into a limit cycle, and postMV reaches the ceiling while UTR and DER read exactly zero.

Following the primary ranking key of this work, the operating point is set by postMV, and the remaining metrics serve as guardrails rather than as selection criteria.
The postMV minimum is flat over 1e-4 to 1e-3 at $E = 16$ and $E = 32$, and over 1e-4 to 3e-4 at $E = 64$, with differences inside these ranges smaller than the seed dispersion.
The plateau shifts left as the expert count grows.
The values carried into the main matrix are its per-scale minima, 1e-3 at $E = 16$, 3e-4 at $E = 32$, and 1e-4 at $E = 64$.
Each doubling of $E$ moves the minimum left by one grid step, a factor of about three, giving a practical rule for transferring $K_I$ across widths without a fresh sweep.
The setpoint $r$ is identical at all three scales by construction, so the shift is not a consequence of per-expert load scale.
We report this rule as an empirical observation on three scales.

Perplexity is flat across the healthy band and shows no trend in $K_I$ there.
Outside the band it degrades in the same direction at all three scales, though the collapsed runs are flagged \textsc{saturated} and enter no comparison.

Saturation is a distinct failure and not merely a large value of postMV.
MaxVio reports only the most loaded expert, so within a single optimization step it attains its ceiling $E/K - 1$ as soon as one expert is selected by every token, whatever the remaining $K - 1$ selections per token do.
Reaching the ceiling therefore constrains only one expert per step, and it says nothing about which expert this is.
Over the observation window this identity shifts, which is why DER and UTR read zero at these settings.
We also report the same pattern in \Cref{app:knob}.

\subsection{Routing sparsity: Top-K ablation}
\label{app:topk}

The pretraining experiment (\Cref{subsec:pretraining}) uses the number of experts $E \in \{16, 32, 64\}$ at a fixed routing sparsity $K/E = 1/8$, where $K$ is the number of experts selected per token under $\operatorname{Top}\text{-}K$ routing.
We further provide an ablation on this routing sparsity with $E=64$ and $K \in \{4, 8, 16\}$ on the \emph{1 $\rightarrow$ 1} scenario, whose configuration is given in \Cref{apptab:topk}.
$K_I=$1e-4 is applied uniformly in this experiment, and other settings remain identical to those described in \Cref{app:setup}.

Moving $\operatorname{Top}\text{-}K$ at fixed $E$ is not free.
\Cref{equ:MaxVio} bounds MaxVio above by the saturation ceiling $\mathrm{MaxVio}^{\mathrm{sat}} = E/K - 1$.
In the pretraining experiments, $K/E$ is fixed at $1/8$ and $\mathrm{MaxVio}^{\mathrm{sat}} = 7$ at every scale.
This ablation changes $K$ and therefore raw MaxVio is not directly comparable across $\operatorname{Top}\text{-}K$.

\begin{figure}[t]
    \begin{center}
    \includegraphics[width=0.99\textwidth]{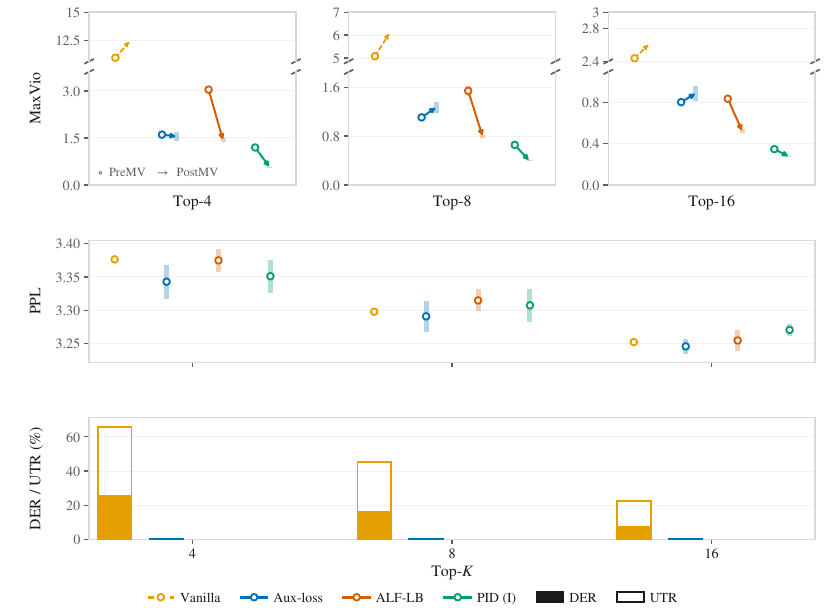}
    \end{center}
    \caption{$\operatorname{Top}\text{-}K$ ablation at $E = 64$ under the \emph{1 $\rightarrow$ 1} scenario.
    Lower is better in every panel.
    ALF-LB and PID (I) have zero DER and UTR at every $K$ and are therefore invisible in the bottom panel.}
    \label{appfig:topk}
\end{figure}

\Cref{appfig:topk} illustrates the ablation result.
The MaxVio ranking is identical at every $\operatorname{Top}\text{-}K$, with PID (I) lowest, followed by ALF-LB, Aux-loss, and Vanilla.
The Aux-loss is the strongest method on perplexity in every column, but its balance degrades across the shift.
Both bias-based balancers, by contrast, absorb the shift and settle below where they entered it.
Neither bias-based balancer registers any dead or under-trained expert at any $\operatorname{Top}\text{-}K$.
These results reproduce the pattern observed in the pretraining experiment at a fixed routing sparsity.
 
\begin{table}[t]
\caption{Configuration of the routing-sparsity ablation on OLMoE-64.}
\label{apptab:topk}
\begin{center}
\begin{tabular}{lccc}
\toprule
$\operatorname{Top}\text{-}K$ & $\operatorname{Top}\text{-}4$ & $\operatorname{Top}\text{-}8$ & $\operatorname{Top}\text{-}16$ \\
\midrule
Routing sparsity $K/E$ & $1/16$ & $1/8$ & $1/4$ \\
Setpoint $r$ & 2048 & 4096 & 8192 \\
Ceiling $\mathrm{MaxVio}^{\mathrm{sat}} = E/K - 1$ & 15 & 7 & 3 \\
\bottomrule
\end{tabular}
\end{center}
\end{table}

\subsection{How Proportional / Integral / Derivative term works}
\label{app:pd}

Taking the difference of \Cref{equ:DiscretePID} between consecutive steps gives the increment each term adds to the bias, the incremental form of PID \citep{franklin2002feedback}
\begin{equation}
\label{equ:Increment}
    \Delta u_i(t) = K_I\, e_i(t) + K_P\, \Delta e_i(t) + K_D\, \Delta^2 e_i(t),
\end{equation}
where $\Delta e_i(t) = e_i(t) - e_i(t-1)$ and $\Delta^2 e_i(t) = \Delta e_i(t) - \Delta e_i(t-1)$.
Under this form the integral term moves the bias by a copy of the error, the proportional term by its first difference and the derivative term by its second difference.

Here we verify the statements of \Cref{subsec:terms} on the configuration with $E=64$ and $\operatorname{Top}\text{-}8$ under the \emph{1 $\rightarrow$ 1} scenario of \Cref{subsec:pretraining}.
Two baselines are used throughout.
One is the router without any balancing mechanism, written Vanilla.
The other is the integral balancer at $K_I=$ 1e-4, which the rest of this section refers to as PID (I).
The proportional term is run alone at five gains from 1e-4 to 1 spaced by a factor of ten, and the derivative term alone at three gains from 1e-4 to 1e-2 with the same spacing.
In every such run the other two gains are zero.
The comparison here made on the load balance metrics alone.

PreMV, PostMV and DER follow \Cref{sec:experiment} and three metrics are added for this study.
\emph{RMS$|u|$} is the root mean square of the bias over experts, layers and steps in the PostMV window and measures how hard the balancer pushes on the router.
Since the bias is added directly to the softmax scores in \Cref{equ:Gate}, it lives on the same scale as those scores.
A larger RMS$|u|$ means the balancer overrides the router's own preferences more strongly.
\emph{Flip} is the fraction of consecutive bias increments $u_i(t) - u_i(t-1)$ that change sign in the same window.
By \Cref{equ:Increment} an increment is zero when the error, its first difference or its second difference is zero for the corresponding term, which for the proportional and derivative terms happens most often when an expert receives no tokens on consecutive steps.
Such an increment has no sign, and every consecutive pair in which either increment is zero is therefore left out of the count.
\emph{Active} is the fraction of experts that receive at least one token in a step, averaged over steps and layers in the same window.
DER counts an expert whose load averaged over the window falls below $0.01\,r$, and the average hides whether the expert was starved on every step or only on some of them.
An expert that is excluded on some steps and included on others lowers Active while leaving DER unchanged, whereas an expert that stays excluded for the whole window lowers Active and raises DER.
Read together, PostMV is the outcome and RMS$|u|$ is the effort spent on it.
Flip tells whether that effort is sustained or reverses from step to step, and Active with DER tells whether experts are lost intermittently or persistently.
A large RMS$|u|$ with a low Flip is a balancer holding a steady bias.
A large RMS$|u|$ with a high Flip is a balancer pushing hard in alternating directions, and most of the effort then cancels while PostMV stays poor.
\Cref{tab:pd:iso} reports above metrics for every cell.

\begin{table}[t]
\caption{Isolated effects of the P, I and D gains.}
\label{tab:pd:iso}
\begin{center}
\newcommand{\ctr}[1]{\multicolumn{1}{c}{#1}}
\begin{tabular}{llrrrrrr}
\toprule
Method & Gains & PreMV $\downarrow$ & PostMV $\downarrow$ & RMS$|u|$ & DER $\downarrow$ & Active  $\uparrow$ & Flip \\
\midrule
Vanilla & \ctr{--} & 5.08 & 6.08 & \ctr{--} & 16.3\% & 94.3\% & \ctr{--} \\
PID (I) & $K_I=$ 1e-4 & 0.62 & 0.40 & 4.0e-2 & 0.0\% & 100.0\% & 37.6\% \\
\midrule
PID (P) & $K_P=$ 1e-4 & 5.17 & 5.62 & 1.6e-4 & 15.0\% & 92.8\% & 67.7\% \\
PID (P) & $K_P=$ 1e-3 & 4.35 & 5.10 & 1.4e-3 & 10.7\% & 96.8\% & 69.2\% \\
PID (P) & $K_P=$ 1e-2 & 2.72 & 2.86 & 8.6e-3 & 1.2\% & 98.2\% & 84.5\% \\
PID (P) & $K_P=$ 1e-1 & 5.22 & 5.33 & 1.6e-1 & 5.6\% & 79.1\% & 94.6\% \\
PID (P) & $K_P=$ 1 & 5.62 & 5.73 & 1.8 & 8.3\% & 58.0\% & 97.1\% \\
\midrule
PID (D) & $K_D=$ 1e-4 & 5.08 & 5.67 & 1.3e-5 & 15.3\% & 94.1\% & 74.1\% \\
PID (D) & $K_D=$ 1e-3 & 5.12 & 5.47 & 4.5e-4 & 16.1\% & 93.4\% & 76.1\% \\
PID (D) & $K_D=$ 1e-2 & 6.98 & 6.99 & 3.3e-2 & 37.3\% & 82.2\% & 84.9\% \\
\bottomrule
\end{tabular}
\end{center}
\end{table}

\subsubsection{The error spectrum}
\label{app:pd:spectra}

\begin{figure}[t]
\centering
\includegraphics[width=0.9\linewidth]{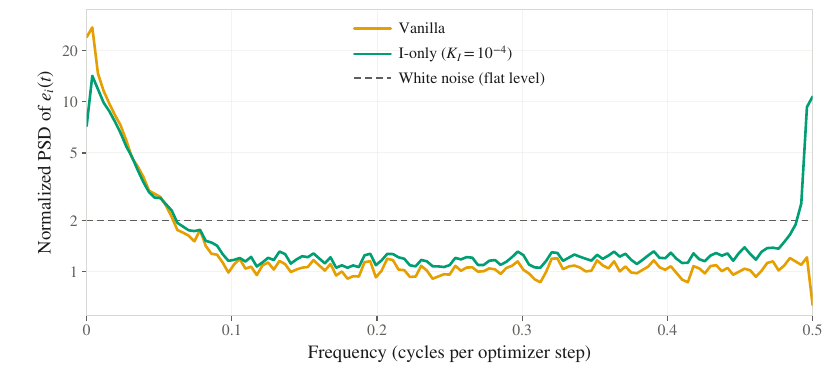}
\caption{Per-step error spectrum in the PostMV window for Vanilla and PID (I).
Welch estimate with a Hann window of 256 steps and half overlap, no per-segment detrend, bin width $1/256$ cycles per step.
The estimate is one-sided, folding the negative frequencies of the real-valued sequence onto $[0, 0.5]$, and the endpoint bins $f = 0$ and $f = 0.5$ sit at half the interior density by construction.}
\label{fig:pd:F1}
\end{figure}

\Cref{fig:pd:F1} shows the power spectrum of the per-step error $e_i(t)$ over the PostMV window.
Each (layer, expert) series has its mean removed before the estimate, the $16 \times 64$ estimates are combined by a power-weighted mean, and the result is normalized to unit total power.
Frequency is measured in cycles per optimizer step, the inverse of the period in steps.
Power at $f = 0.01$ therefore belongs to an expert whose load swings above and below its share with a period of 100 steps, staying overloaded for about 50 consecutive steps and then underloaded for the next 50.
Such an imbalance outlives many batches.
Power at $f = 0.5$ belongs to the opposite pattern, an expert overloaded on one step and underloaded on the next, the fastest alternation a per-step sequence can hold.
Under this normalization a white noise sits at the flat level $1/0.5 = 2$ over the whole band.

Vanilla and PID (I) share the same shape over most of the band.
Both have a slow component near zero frequency, where the Vanilla density reaches 27.5 at the lowest interior bin ($f \approx 0.004$), and a flat floor of about 1 from $f \approx 0.1$ onward.
Integrating the spectrum puts 25.9\% of the Vanilla error variance in components slower than 100 steps ($f \le 0.01$), against 13.0\% for PID (I) and 1.9\% for white noise through the same estimator.
PID (I) changes the spectrum in two places only.
It halves the slow component, the part of the imbalance an integrator can follow.
It also adds a narrow peak at the alternation frequency $f = 0.5$, where its density is 10.8 against 0.63 for Vanilla in the endpoint bin.
This peak is the alternating error of \Cref{subsec:terms} in its measured form.
Because the bias acts one step late, an expert near a selection tie is pushed in on one step and pushed out on the next.
At $K_I =$ 1e-4 this alternation occupies only the top of the band and leaves the rest of the spectrum unchanged.

If every token chose its $K$ experts independently and uniformly, $L_i(t)$ would be binomial with variance $r(1 - K/E)$ and a standard deviation of about 60 tokens around $r = 4096$ here.
The measured per-step standard deviation of $L_i(t)$ over the PostMV window is about 300 tokens at the median (layer, expert) series, five times the binomial value, and it varies by four orders of magnitude across series.
Batch composition rather than sampling noise therefore dominates the error, and the flat floor with it.
 
\subsubsection{Spectral signature of each term}
\label{app:pd:signature}

By \Cref{equ:Increment} the integral term moves the bias by a copy of the error, the proportional term by its first difference and the derivative term by its second difference.
Each difference multiplies the density at frequency $f$ by $4\sin^2(\pi f)$, which is zero at $f = 0$ and 4 at $f = 0.5$.
The increment spectrum of the proportional term is therefore the error spectrum with its slow components suppressed and its alternation amplified, and that of the derivative term applies the same reweighting twice.

\begin{figure}[t]
\centering
\includegraphics[width=0.9\linewidth]{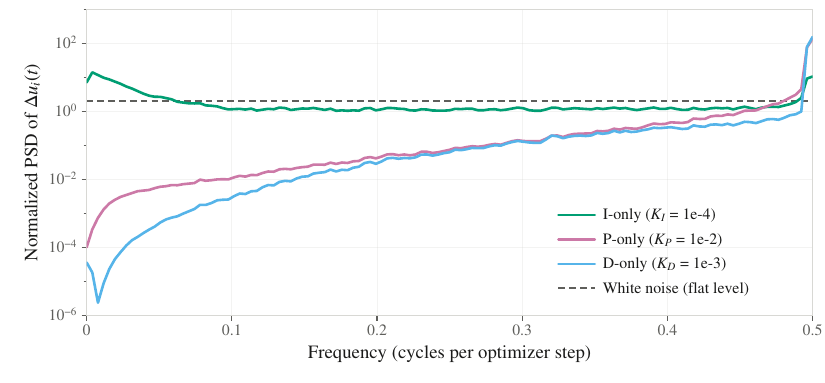}
\caption{Increment spectrum of $\Delta u_i(t)$ over the PostMV window, same estimator and normalization as \Cref{fig:pd:F1}.
The estimate is one-sided, so the endpoint bins $f = 0$ and $f = 0.5$ sit at half the interior density by construction.}
\label{fig:pd:F2}
\end{figure}

\Cref{fig:pd:F2} shows the spectrum of the bias increments $\Delta u_i(t)$ over the PostMV window for PID (I) and for the proportional and derivative cells with the lowest PostMV in \Cref{tab:pd:iso}, $K_P =$ 1e-2 and $K_D =$ 1e-3.
Under the integral term the increment is $K_I e_i(t)$ and its spectrum reproduces \Cref{fig:pd:F1}, with 0.383 of its power above $f = 0.25$.
Under the proportional term the increment spectrum rises from about 1e-4 near zero frequency to above 1e2 at $f = 0.5$, with 0.994 of the power above $f = 0.25$.
Under the derivative term it starts below the proportional curve and reaches the same level at $f = 0.5$, with 0.996 of the power above $f = 0.25$.
For comparison, white noise differenced once and twice places 0.818 and 0.924 of its power there.

The sign-flip rate of the increments compresses the same information into a single number.
White noise differenced once changes sign between consecutive steps with probability $0.667$ and differenced twice with probability $0.732$ \citep{kedem1986zero}.
These are the rates the proportional and derivative increments approach at small gain, since both differences discard the slow component of the error and leave a sequence close to white.
The integral increment is a copy of the error and changes sign only as often as the error itself does, which is below one half for an error dominated by its slow component.
\Cref{tab:pd:iso} confirms this pattern in its Flip column.
At the smallest gains both memoryless terms flip at the white-noise rates, 67.7\% for the proportional term and 74.1\% for the derivative term.
Raising $K_P$ makes the bias large enough to change the selection at every step and the proportional rate climbs to 97.1\%, whereas the derivative rate moves only to 76.1\% before the run collapses.
PID (I) flips 37.6\% of the time, below one half as expected for an error dominated by its slow component.
Only the integral increment follows the error while the other two follow its alternation as \Cref{subsec:terms} mentioned.

\subsubsection{The proportional term alone}
\label{app:pd:p}

Reading \Cref{tab:pd:iso} down the PID (P) rows shows that the proportional term alone does balance.
At $K_P = $ 1e-2 PostMV is 2.86 against 6.08 for Vanilla and DER is 1.2\% against 16.3\%.
Ten times smaller, at $K_P = $ 1e-3, the gain leaves PostMV at 5.10 and close to Vanilla.
Ten times larger, at $K_P = $ 1e-1, it brings PostMV back to 5.33.
The two sides fail for different reasons.
A dead expert has $e_i = 1$ at every step and receives from the proportional term a constant bias of $K_P$.
This bias pulls the expert back into selection only if $K_P$ alone exceeds the score gap that keeps it out.
Accordingly DER falls from 15.0\% to 10.7\% to 1.2\% as the gain grows and more experts cross their gap.
At large gain the bias is strong enough to change the selection at every step and the feedback itself becomes the problem.
Because the bias computed after step $t$ acts at step $t+1$, a term without memory over-corrects and reverses on alternate steps.
Flip climbs from 84.5\% to 97.1\% over the three largest gains, with nearly every increment undoing the previous one.
Active drops to 79.1\% and 58.0\% at the two largest gains, since experts excluded on one step are included again on the next.
The improvement at the optimum is already present before the distribution shift, with PreMV 2.72 against 5.08 for Vanilla.
Even at its best the proportional term leaves PostMV at about 7 times the value reached by PID (I) and DER at 1.2\% against zero.
A constant offset reduces a persistent error by a finite factor but cannot remove it, and only the ramp of the integral term removes the residual.

\subsubsection{The derivative term alone}
\label{app:pd:d}

At $K_D = $ 1e-4 and 1e-3 the derivative term leaves PostMV at 5.67 and 5.47 against 6.08 for Vanilla.
DER stays at 15.3\% and 16.1\% against 16.3\%, and RMS$|u|$ remains below 5e-4.
The term thus leaves the dead experts untouched and only slightly lowers PostMV.
A persistent error does not change from one step to the next, and a dead expert or a lasting imbalance therefore produces no output from the term.
What the term acts on is the second difference of the fluctuating part of the error alone.
Flip confirms this at 74.1\% and 76.1\%, close to the 73.2\% of the second difference of white noise.
At $K_D = $ 1e-2 the run collapses before the distribution shift, with PreMV already at 6.98 against the ceiling of 7.
Once an expert receives no tokens its error stops changing and the term falls silent on it, and the experts excluded at this gain stay excluded.
Active accordingly falls to 82.2\% against 94.3\% for Vanilla while DER rises to 37.3\% against 16.3\%, the signature of experts lost persistently rather than intermittently.
No useful range for the derivative term was observed on the tested grid in this configuration.
Below 1e-2 its output is too small to change the balance and at 1e-2 it removes experts from the routing permanently.

\subsubsection{Combinations with the integral term}
\label{app:pd:combo}

The previous two subsections examined the proportional and derivative terms on their own.
Neither removes a persistent error, and both follow the fast part of the error rather than its slow part.
This section adds each of them to PID (I) at the same $K_I = \text{1e-4}$ and asks whether they help the integrator at all.
\Cref{subsec:terms} states that next to the integral term the other two add nothing on the persistent part and only their swing on the fast part.
\Cref{tab:pd:combo} reports the same metrics as \Cref{tab:pd:iso} for every combination.
These metrics describe the closed loop as a whole and cannot attribute its behavior to individual terms.
\Cref{tab:pd:decomp} therefore separates the bias into its three components and splits each into its mean over the PostMV window and its fluctuation about that mean.

\begin{table}[t]
\caption{Combination effects of P, I, and D gains at $K_I = \text{1e-4}$.}
\label{tab:pd:combo}
\begin{center}
\resizebox{0.99\textwidth}{!}{%
\begin{tabular}{llrrrrrr}
\toprule
Method & Gains & PreMV $\downarrow$ & PostMV $\downarrow$ & RMS$|u|$ & DER $\downarrow$ & Active $\uparrow$ & Flip \\
\midrule
PID (I) & $K_I=\text{1e-4}$ & 0.62 & 0.40 & 4.0e-2 & 0.0\% & 100.0\% & 37.6\% \\
\midrule
PID (PI) & $K_P=\text{1e-4}$ & 0.81 & 0.46 & 4.1e-2 & 0.0\% & 100.0\% & 58.6\% \\
PID (PI) & $K_P=\text{1e-3}$ & 1.23 & 0.62 & 3.4e-2 & 0.0\% & 99.9\% & 70.2\% \\
PID (PI) & $K_P=\text{1e-2}$ & 2.79 & 1.84 & 3.2e-2 & 0.0\% & 98.7\% & 83.6\% \\
\midrule
PID (DI) & $K_D=\text{1e-4}$ & 1.06 & 0.52 & 4.1e-2 & 0.0\% & 100.0\% & 69.1\% \\
PID (DI) & $K_D=\text{1e-3}$ & 2.58 & 1.47 & 3.9e-2 & 0.0\% & 99.9\% & 77.9\% \\
\midrule
PID (PID) & $K_P=\text{1e-3}, K_D=\text{1e-3}$ & 2.78 & 1.89 & 4.2e-2 & 0.0\% & 99.5\% & 77.5\% \\
PID (PID) & $K_P=\text{1e-2}, K_D=\text{1e-3}$ & 4.42 & 3.30 & 4.4e-2 & 0.0\% & 98.0\% & 84.0\% \\
\bottomrule
\end{tabular}%
}
\end{center}
\end{table}

\begin{table}[t]
\caption{Contribution of each term to the bias in the PostMV window, split into its mean and its fluctuation about that mean, both as RMS over experts and layers. $\rho$ is the RMS of the proportional and derivative increments over that of the integral increment. A dash marks a term with zero gain.}
\label{tab:pd:decomp}
\begin{center}
\resizebox{0.85\textwidth}{!}{%
\setlength{\tabcolsep}{4pt}
\newcommand{\ctr}[1]{\multicolumn{1}{c}{#1}}
\begin{tabular}{llrrrrrrr}
\toprule
& & \multicolumn{3}{c}{mean} & \multicolumn{3}{c}{fluctuation} & \\
\cmidrule(lr){3-5} \cmidrule(lr){6-8}
Method & Gains & I & P & D & I & P & D & $\rho$ \\
\midrule
PID (I)   & $K_I=\text{1e-4}$ & 4.0e-2 & \ctr{--} & \ctr{--} & 7.8e-4 & \ctr{--} & \ctr{--} & \ctr{0} \\
PID (P)   & $K_P=\text{1e-2}$ & \ctr{--} & 7.9e-3 & \ctr{--} & \ctr{--} & 3.6e-3 & \ctr{--} & \ctr{--} \\
PID (D)   & $K_D=\text{1e-3}$ & \ctr{--} & \ctr{--} & 1.3e-7 & \ctr{--} & \ctr{--} & 4.5e-4 & \ctr{--} \\
\midrule
PID (PI)  & $K_P=\text{1e-4}$ & 4.1e-2 & 1.4e-6 & \ctr{--} & 9.3e-4 & 1.8e-5 & \ctr{--} & 1.5 \\
PID (PI)  & $K_P=\text{1e-3}$ & 3.4e-2 & 9.3e-6 & \ctr{--} & 7.4e-4 & 2.5e-4 & \ctr{--} & 17.5 \\
PID (PI)  & $K_P=\text{1e-2}$ & 3.2e-2 & 1.3e-4 & \ctr{--} & 8.3e-4 & 5.9e-3 & \ctr{--} & 189 \\
\midrule
PID (DI)  & $K_D=\text{1e-4}$ & 4.1e-2 & \ctr{--} & 1.6e-8 & 7.5e-4 & \ctr{--} & 3.0e-5 & 3.0 \\
PID (DI)  & $K_D=\text{1e-3}$ & 3.9e-2 & \ctr{--} & 3.8e-7 & 1.2e-3 & \ctr{--} & 8.1e-4 & 32.5 \\
\midrule
PID (PID) & $K_P=\text{1e-3},\,K_D=\text{1e-3}$ & 4.2e-2 & 1.3e-5 & 4.5e-7 & 9.2e-4 & 6.3e-4 & 1.0e-3 & 35.8 \\
PID (PID) & $K_P=\text{1e-2},\,K_D=\text{1e-3}$ & 4.2e-2 & 9.1e-5 & 5.8e-7 & 8.7e-4 & 8.8e-3 & 1.5e-3 & 171 \\
\bottomrule
\end{tabular}%
}
\end{center}
\end{table}

Every combination improves on the proportional or derivative term alone at the same gain, leaves DER at zero and keeps Active above 98\%.
The integral term does its work whichever term is added to it.
However, none of the seven combinations improves on PID (I).
PostMV rises monotonically with the added gain within each block, from 0.46 to 1.84 for PI and from 0.52 to 1.47 for DI.
PreMV shows the same trend, and the degradation is already present before the distribution shift.

Only the part of the bias that persists over many steps can cancel a persistent imbalance.
The mean column in \Cref{tab:pd:decomp} reports that persistent part for each term.
In every combination the integral term provides almost all of it, at 3.2e-2 to 4.2e-2 against 4.0e-2 in PID (I).
When the proportional term works alone it balances through exactly such an offset, and at $K_P = \text{1e-2}$ its output is mostly mean.
Next to the integrator the same term at the same gain loses that offset, from 7.9e-3 to 1.3e-4, because the integrator has already driven the mean error close to zero.
What remains of the proportional output is fluctuation, with a mean of 1.3e-4 against a fluctuation of 5.9e-3 in that cell.
Its values stay at the 5.9e-3 level on most steps but alternate in sign, and average out to a net offset of only 1.3e-4.
The derivative term shows no such change of role, since its mean stays below 1e-6 alone and in every combination.
Once the integrator is present, adding the proportional or derivative term increases the measured fluctuation and postMV.

$\rho$ in \Cref{tab:pd:decomp} measures that fluctuation as the RMS increment of the two terms relative to that of the integrator.
It exceeds 1 in every combination, meaning that even at the smallest gains the added term moves the bias more than the integrator does over the window.
Within each block it grows with the gain, reaching 189 for PI and 32.5 for DI.

Flip and $\rho$ move together.
At $\rho = 1.5$, the smallest proportional gain, Flip already rises from 37.6\% to 58.6\% while PostMV moves only from 0.40 to 0.46.
Flip reaches 83.6\% at $\rho = 189$.
It indicates that a memoryless term is present and its increment is comparable to the integrator's.

The full PID settings do not recover what PI and DI lose.
Each is worse than both of its parents, the PI cell at the same $K_P$ and the DI cell at the same $K_D$.
Neither term therefore helps the integrator at any of the gains tested.

\subsection{Aux-loss coefficient sweep in continual pretraining}
\label{app:p3auxloss}

Aux-loss improves MaxVio only modestly relative to Vanilla in \Cref{tab:cpt}, which raises the question of whether its coefficient $\alpha$ was set too small.
The value used there, $\alpha = 0.01$, is the coefficient OLMoE was pretrained with \citep{muennighoff2025olmoe} and is therefore the natural default for continual pretraining of that checkpoint.
\Cref{apptab:auxalphasweep} sweeps larger values of $\alpha$ under both the cold and the warm setting.
Increasing $\alpha$ improves balance monotonically while PPL and downstream accuracy degrade in step, which matches the trend reported in \Cref{app:knob}.
Such a trade-off is expected whenever the balancing loss is added directly to and competes with the task loss, as in Aux-loss.

\begin{table}[h]
\caption{Aux-loss balance coefficient sweep.}
\label{apptab:auxalphasweep}
\begin{center}
\resizebox{\textwidth}{!}{%
\begin{tabular}{llrrrrrrrrrrr}
\toprule
$\alpha$ & Setting & preMV$\downarrow$ & postMV$\downarrow$ & DER$\downarrow$ / UTR$\downarrow$ & PPL-CPT$\downarrow$ & PPL-Annealing$\downarrow$ & MMLU$\uparrow$ & HellaSwag$\uparrow$ & ARC-C$\uparrow$ & HumanEval$\uparrow$ & MBPP$\uparrow$ \\
\midrule
0.01 & cold & - & 4.43 & 0.00\% / 1.56\% & 2.19 & 7.21 & 0.49 & 0.74 & 0.50 & 0.12 & 0.18 \\
0.08 & cold & - & 2.30 & 0.00\% / 0.00\% & 2.22 & 7.32 & 0.46 & 0.73 & 0.50 & 0.11 & 0.20 \\
0.4  & cold & - & 0.95 & 0.00\% / 0.00\% & 2.25 & 7.60 & 0.46 & 0.72 & 0.49 & 0.11 & 0.14 \\
0.8  & cold & - & 0.60 & 0.00\% / 0.00\% & 2.27 & 7.77 & 0.42 & 0.71 & 0.51 & 0.11 & 0.14 \\
4    & cold & - & 0.23 & 0.00\% / 0.00\% & 2.29 & 8.38 & 0.41 & 0.71 & 0.51 & 0.10 & 0.13 \\
\midrule
\textit{PID (I)} & cold & - & 0.90 & 0.00\% / 0.00\% & 2.19 & 7.39 & 0.46 & 0.72 & 0.50 & 0.12 & 0.14 \\
\midrule
0.01 & warm & 1.09 & 4.51 & 0.00\% / 6.93\% & 2.24 & 6.07 & 0.51 & 0.77 & 0.56 & 0.13 & 0.19 \\
0.08 & warm & 0.89 & 2.53 & 0.00\% / 0.10\% & 2.27 & 6.26 & 0.49 & 0.76 & 0.55 & 0.11 & 0.14 \\
0.4  & warm & 0.61 & 1.11 & 0.00\% / 0.00\% & 2.31 & 6.33 & 0.46 & 0.75 & 0.54 & 0.12 & 0.13 \\
0.8  & warm & 0.47 & 0.71 & 0.00\% / 0.00\% & 2.34 & 6.43 & 0.46 & 0.75 & 0.55 & 0.10 & 0.12 \\
4    & warm & 0.20 & 0.27 & 0.00\% / 0.00\% & 2.38 & 6.67 & 0.42 & 0.74 & 0.55 & 0.10 & 0.12 \\
\midrule
\textit{PID (I)} & warm & 0.66 & 0.87 & 0.00\% / 0.00\% & 2.23 & 6.28 & 0.48 & 0.75 & 0.55 & 0.11 & 0.13 \\
\bottomrule
\end{tabular}%
}
\end{center}
\end{table}

\subsection{Continual pretraining at longer budgets}
\label{app:cptmore}

The continual pretraining (CPT) results in \Cref{subsec:continualpretraining} use a budget of 16,500 steps on the code corpus.
This subsection repeats the cold and warm start settings at two and five times that budget (33,000 and 82,500 CPT steps), which we call $2\times$ and $5\times$ below.
In the warm start setting the annealing segment is unchanged at 16,500 steps on the same corpus, and only the CPT budget varies.
All other settings are identical to \Cref{subsec:continualpretraining}, including the steady-state window length, the held-out sets and the evaluation pipeline.

\paragraph{Balance.}
\Cref{apptab:cptmore_metrics} reports preMV, postMV, PPL-CPT and PPL-Annealing at each budget.
The ordering of postMV in every setting and at every budget is the same as at $1\times$, namely PID (I) $<$ ALF-LB $<$ Aux-loss $<$ Vanilla.
The gap between PID (I) and ALF-LB does not close with budget.
It stays flat in the cold start setting and widens slightly in the warm start setting.
Both bias-based balancers report zero dead and zero under-trained experts in every setting at every budget (\Cref{apptab:cptmore_metrics_2}).
Aux-loss eliminates dead experts but leaves a residual fraction under-trained.
Vanilla leaves experts both dead and under-trained throughout, and neither metric improves as the budget grows.

\paragraph{Language modeling.}
PPL-CPT keeps falling with budget for every method and has not reached a plateau at $5\times$.
In the cold start setting the bias balancers sit slightly above Vanilla and Aux-loss on PPL-CPT at the longer budgets.
The warm start setting shows no such separation, with the four methods within 0.02 of each other at every budget.
PPL-Annealing grows with budget for every method, but not uniformly.
The extra forgetting of the bias balancers relative to Vanilla and Aux-loss is already present in the $2\times$ runs and does not keep widening at $5\times$.
PID (I) and ALF-LB are indistinguishable on both perplexities at every budget.

\begin{table}[t]
\caption{preMV, postMV, PPL-CPT and PPL-Annealing at three CPT budgets. preMV is defined only in the warm start setting, where the annealing corpus and steps are unchanged across budgets and the values are therefore identical.}
\label{apptab:cptmore_metrics}
\begin{center}
\resizebox{0.99\textwidth}{!}{%
\begin{tabular}{llccc ccc ccc ccc}
\toprule
& & \multicolumn{3}{c}{preMV $\downarrow$} & \multicolumn{3}{c}{postMV $\downarrow$} & \multicolumn{3}{c}{PPL-CPT $\downarrow$} & \multicolumn{3}{c}{PPL-Annealing $\downarrow$} \\
\cmidrule(lr){3-5}\cmidrule(lr){6-8}\cmidrule(lr){9-11}\cmidrule(lr){12-14}
Setting & Method & $1\times$ & $2\times$ & $5\times$ & $1\times$ & $2\times$ & $5\times$ & $1\times$ & $2\times$ & $5\times$ & $1\times$ & $2\times$ & $5\times$ \\
\midrule
cold & Vanilla  & -- & -- & -- & 5.83 & 5.82 & 5.76 & 2.19 & 2.11 & 2.02 & 7.23 & 7.33 & 7.53 \\
     & Aux-loss & -- & -- & -- & 4.43 & 4.22 & 4.04 & 2.19 & 2.11 & 2.02 & 7.21 & 7.34 & 7.53 \\
     & ALF-LB   & -- & -- & -- & 1.01 & 1.05 & 0.97 & 2.19 & 2.14 & 2.05 & 7.36 & 7.98 & 8.18 \\
     & PID (I)  & -- & -- & -- & 0.90 & 0.94 & 0.85 & 2.19 & 2.13 & 2.03 & 7.39 & 7.98 & 8.19 \\
\midrule
warm & Vanilla  & 1.14 & 1.14 & 1.14 & 5.76 & 5.77 & 5.79 & 2.24 & 2.18 & 2.10 & 6.07 & 6.15 & 6.28 \\
     & Aux-loss & 1.09 & 1.09 & 1.09 & 4.51 & 4.39 & 4.28 & 2.24 & 2.18 & 2.11 & 6.07 & 6.15 & 6.28 \\
     & ALF-LB   & 0.72 & 0.72 & 0.72 & 1.03 & 1.06 & 0.98 & 2.24 & 2.18 & 2.12 & 6.28 & 6.58 & 6.75 \\
     & PID (I)  & 0.66 & 0.66 & 0.66 & 0.87 & 0.90 & 0.79 & 2.23 & 2.18 & 2.11 & 6.28 & 6.58 & 6.75 \\
\bottomrule
\end{tabular}%
}
\end{center}
\end{table}

\begin{table}[t]
\caption{DER and UTR at three CPT budgets}
\label{apptab:cptmore_metrics_2}
\begin{center}
\resizebox{0.8\textwidth}{!}{%
\begin{tabular}{llccc ccc}
\toprule
& & \multicolumn{3}{c}{DER $\downarrow$} & \multicolumn{3}{c}{UTR $\downarrow$} \\
\cmidrule(lr){3-5}\cmidrule(lr){6-8}
Setting & Method & $1\times$ & $2\times$ & $5\times$ & $1\times$ & $2\times$ & $5\times$ \\
\midrule
cold & Vanilla  & 1.07\% & 1.07\% & 1.27\% & 29.10\% & 29.30\% & 29.30\% \\
     & Aux-loss & 0.00\% & 0.00\% & 0.00\% & 1.56\% & 1.27\% & 0.59\% \\
     & ALF-LB   & 0.00\% & 0.00\% & 0.00\% & 0.00\% & 0.00\% & 0.00\% \\
     & PID (I)  & 0.00\% & 0.00\% & 0.00\% & 0.00\% & 0.00\% & 0.00\% \\
\midrule
warm & Vanilla  & 0.59\% & 0.78\% & 0.68\% & 27.73\% & 29.00\% & 31.15\% \\
     & Aux-loss & 0.00\% & 0.00\% & 0.00\% & 6.93\% & 4.69\% & 3.71\% \\
     & ALF-LB   & 0.00\% & 0.00\% & 0.00\% & 0.00\% & 0.00\% & 0.00\% \\
     & PID (I)  & 0.00\% & 0.00\% & 0.00\% & 0.00\% & 0.00\% & 0.00\% \\
\bottomrule
\end{tabular}%
}
\end{center}
\end{table}

\paragraph{Benchmarks.}
\Cref{tab:cptmore_bench} reports the five benchmarks at all three budgets, with the $1\times$ columns repeating \Cref{tab:cpt}.
MMLU and HellaSwag drift down slightly with budget for every method, consistent with the forgetting trend above.
ARC-Challenge, HumanEval and MBPP move within their sampling error in both directions.
The ordering between methods is unchanged from $1\times$ on every task.
Vanilla and Aux-loss score higher than the bias balancers on MMLU, HellaSwag and MBPP.
All four methods are indistinguishable on ARC-Challenge and HumanEval, and neither PID (I) nor ALF-LB is consistently better than the other.
Longer CPT therefore does not change the conclusions of \Cref{subsec:continualpretraining}.

\begin{table}[t]
\caption{Benchmarks at three CPT budgets.}
\label{tab:cptmore_bench}
\begin{center}
\resizebox{0.99\textwidth}{!}{%
\begin{tabular}{llrrr rrr rrr rrr rrr}
\toprule
& & \multicolumn{3}{c}{MMLU $\uparrow$} & \multicolumn{3}{c}{HellaSwag $\uparrow$} & \multicolumn{3}{c}{ARC-C $\uparrow$} & \multicolumn{3}{c}{HumanEval $\uparrow$} & \multicolumn{3}{c}{MBPP $\uparrow$} \\
\cmidrule(lr){3-5}\cmidrule(lr){6-8}\cmidrule(lr){9-11}\cmidrule(lr){12-14}\cmidrule(lr){15-17}
Method & Setting & $1\times$ & $2\times$ & $5\times$ & $1\times$ & $2\times$ & $5\times$ & $1\times$ & $2\times$ & $5\times$ & $1\times$ & $2\times$ & $5\times$ & $1\times$ & $2\times$ & $5\times$ \\
\midrule
Vanilla  & cold & 0.50 & 0.50 & 0.48 & 0.74 & 0.73 & 0.73 & 0.50 & 0.49 & 0.51 & 0.12 & 0.14 & 0.14 & 0.18 & 0.22 & 0.19 \\
Aux-loss & cold & 0.49 & 0.50 & 0.48 & 0.74 & 0.73 & 0.73 & 0.50 & 0.50 & 0.51 & 0.12 & 0.12 & 0.14 & 0.18 & 0.21 & 0.18 \\
ALF-LB   & cold & 0.46 & 0.47 & 0.46 & 0.71 & 0.71 & 0.70 & 0.48 & 0.49 & 0.49 & 0.10 & 0.11 & 0.10 & 0.14 & 0.12 & 0.16 \\
PID (I)  & cold & 0.46 & 0.45 & 0.44 & 0.72 & 0.71 & 0.70 & 0.50 & 0.49 & 0.49 & 0.12 & 0.12 & 0.13 & 0.14 & 0.16 & 0.16 \\
\midrule
Vanilla  & warm & 0.51 & 0.51 & 0.50 & 0.77 & 0.76 & 0.76 & 0.55 & 0.55 & 0.55 & 0.12 & 0.12 & 0.10 & 0.18 & 0.20 & 0.20 \\
Aux-loss & warm & 0.51 & 0.51 & 0.50 & 0.77 & 0.76 & 0.76 & 0.56 & 0.56 & 0.55 & 0.13 & 0.15 & 0.10 & 0.19 & 0.17 & 0.18 \\
ALF-LB   & warm & 0.48 & 0.48 & 0.46 & 0.75 & 0.75 & 0.74 & 0.54 & 0.53 & 0.53 & 0.10 & 0.10 & 0.10 & 0.13 & 0.15 & 0.13 \\
PID (I)  & warm & 0.48 & 0.48 & 0.47 & 0.75 & 0.74 & 0.74 & 0.55 & 0.55 & 0.54 & 0.11 & 0.10 & 0.11 & 0.13 & 0.15 & 0.16 \\
\bottomrule
\end{tabular}%
}
\end{center}
\end{table}

\section{Additional discussion} 

\subsection{Specialization probe}
\label{app:specialization}

We quantify how load balance trades against expert specialization in the setting of \Cref{subsec:continualpretraining}.
The balancers act on the router in different places.
Aux-loss adds a loss term whose gradient reshapes the gate weights.
ALF-LB and PID (I) add no term to the training objective and instead shift the gate scores by a bias at selection time.
Here we ask how much input dependence each balancer removes from routing, and whether this reduction is encoded in the gate weights or arises only at expert selection.

\paragraph{Measurement.}
We use the routing mutual information between input domain and selected expert, which captures domain-level specialization.
The probe takes exactly 256 held-out instances of 4096 tokens from each of $D=7$ domains, which comprise the six domains of the annealing corpus and the CPT corpus.
Equal sampling gives $p(d)=1/D$ and fixes the domain entropy at $H(D)=\ln 7\approx 1.95$.
A single inference-only forward pass with the bias held fixed accumulates the (token, slot) assignments of each MoE layer $l$ into integer counts $C[l,e,d]$, where $C[l,e,d]$ is the number of assignments from domain $d$ to expert $e$.
From these counts we estimate
\begin{equation}
I_l = H\!\left(p(e\mid l)\right) - \sum_d p(d)\, H\!\left(p(e\mid l,d)\right),
\end{equation}
which is zero when every domain uses the experts in the same proportions and reaches $H(D)$ only when the selected expert uniquely determines the domain.
We report its mean over the 16 MoE layers.
Randomly permuting the domain labels across instances preserves every routing decision but removes any true dependence between expert and domain, which lowers the estimate to 0.0007 ($0.04\%$ of $\ln 7$).
The base model reads about 7\% of $\ln 7$, far above this floor.
Across 2000 bootstrap resamples of the probe instances the 95\% interval of every layer-mean $I$ is narrower than 0.006.
A collapsed router would also read a low $I_l$ since $I_l \le H(p(e\mid l))$.
However, the routing entropy stays within 3\% of its uniform bound $\ln 64$ in every model we report, which rules out collapse as the cause of any reduction below.
 
To place each model on the balance axis we compute MaxVio from the same pass.
It aggregates all probe tokens of the corpus a setting trains on, the CPT domain at the end of CPT and the annealing domains in their training proportions at the end of the annealing phase.
We call it probe MaxVio.
Aggregated over about a million tokens per domain, it is not comparable in scale to the per-batch MaxVio of the main text.
 
Each ALF-LB and PID (I) model is measured twice.
The deployed reading uses the selection the model actually makes from the gate scores plus the bias.
The bias-removed reading repeats the forward pass with the stored bias set to zero in every layer.
Selection then follows the gate scores alone, and every later layer receives the inputs produced by this selection.
The two readings coincide for Vanilla and Aux-loss because these models carry no bias.

\begin{table}[t]
\caption{Probe MaxVio and layer-mean routing mutual information $I$.
Annealing is the warm start at the end of the annealing phase (step 16,500) with MaxVio over the annealing domains.
The CPT columns are the end of CPT (step 16,500 cold, 33,000 warm) with MaxVio over the CPT domain.
Bias removed repeats the probe with the stored bias set to zero.
}
\label{apptab:specialization}
\begin{center}
\resizebox{0.8\linewidth}{!}{%
\begin{tabular}{l cc cc cc}
\toprule
& \multicolumn{2}{c}{Annealing} & \multicolumn{2}{c}{CPT, cold start} & \multicolumn{2}{c}{CPT, warm start} \\
\cmidrule(lr){2-3}\cmidrule(lr){4-5}\cmidrule(lr){6-7}
Method & MaxVio & $I$ & MaxVio & $I$ & MaxVio & $I$ \\
\midrule
Base                 & 0.83 & 0.141 & 5.23 & 0.141 & 5.23 & 0.141 \\
Vanilla              & 0.91 & 0.146 & 5.79 & 0.137 & 5.71 & 0.143 \\
Aux-loss             & 0.88 & 0.135 & 4.29 & 0.110 & 4.42 & 0.114 \\
\midrule
ALF-LB               & 0.34 & 0.145 & 0.46 & 0.074 & 0.51 & 0.077 \\
\quad bias removed   & 0.87 & 0.147 & 4.14 & 0.110 & 4.61 & 0.125 \\
PID (I)              & 0.07 & 0.145 & 0.11 & 0.074 & 0.11 & 0.077 \\
\quad bias removed   & 0.88 & 0.147 & 4.14 & 0.111 & 4.62 & 0.125 \\
\bottomrule
\end{tabular}%
}
\end{center}
\end{table}
 
\begin{figure}[t]
    \begin{center}
    \includegraphics[width=0.99\textwidth]{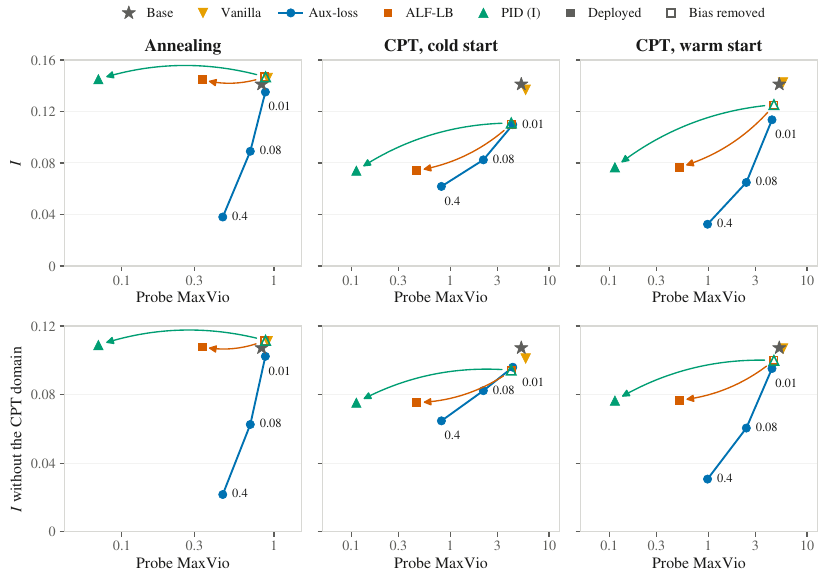}
    \end{center}
    \caption{Routing mutual information $I$ against probe MaxVio on a log axis.
    The top row uses all seven probe domains and the bottom row excludes the CPT domain.
    The line joins the three Aux-loss runs, labeled by coefficient.
    Arrows point from the bias-removed reading to the deployed one.
    Further left is more balanced and higher is more specialized.
    }
    \label{appfig:specialization}
\end{figure}
 
\paragraph{Balance against specialization.}
\Cref{apptab:specialization} and \Cref{appfig:specialization} place every model by its probe MaxVio and its routing mutual information.
Methods are compared on their deployed reading, with the bias applied.
In addition to $\alpha = 0.01$, we further train Aux-loss with $\alpha = 0.08$ and $\alpha = 0.4$ to move its probe MaxVio toward that of the bias-based balancers.
Raising $\alpha$ balances the load further and removes more routing information at every step, in both start settings and on the annealing corpus.
Even at $\alpha = 0.4$ its load stays less balanced than that of either bias-based balancer, and it keeps less routing information than they do.
The bias-based balancers have both lower probe MaxVio and higher domain-routing mutual information than others.
Removing the CPT domain from the probe leaves these conclusions unchanged (bottom row of \Cref{appfig:specialization}).
ALF-LB and PID (I) carry nearly the same routing information in every setting and reading, while PID (I) reaches the lower probe MaxVio.
 
\paragraph{Where the bias costs specialization.}
On the annealing corpus the bias costs almost nothing.
Both bias-based balancers bring probe MaxVio well below that of Vanilla while keeping nearly the same routing information.
Unlike the annealing corpus, the CPT corpus consists of a single domain.
Balancing a single-domain stream means spreading that domain evenly over the experts, and the bias does exactly this.
About half of the deployed reduction comes from the CPT domain itself.
The other half comes from the annealing domains, which the same bias pushes toward a more similar use of the experts.
Removing the bias gives back more than half of the deployed reduction.
What remains has been learned into the weights, and it is larger in the cold start, where the balancer starts from zero at the distribution shift instead of running through the annealing phase.
At a similar probe MaxVio, the weights trained with the bias carry about as much routing information as Aux-loss at 0.01 in the cold start and more in the warm start.
Aux-loss, as expected from its mechanism, pays its entire cost in the weights.
 
\paragraph{Removing the bias at inference.}
To measure how removing the bias affects perplexity, we take the four ALF-LB and PID (I) models at the end of CPT, set their bias to zero, and re-evaluate them with everything else unchanged.
Perplexity falls for both bias-based methods on both sets in \Cref{apptab:zero_bias_ppl}.
Balance does not survive the removal because the load on the CPT domain in the probe returns most of the way to the imbalance of Vanilla.

\begin{table}[t]
\caption{Held-out perplexity with the routing bias applied (on) and set to zero (off).
$\Delta$ is $\mathrm{PPL}_{\text{off}}/\mathrm{PPL}_{\text{on}}-1$.
The base is the pre-trained model before continual pretraining.
``--'' marks methods without a routing bias.}
\label{apptab:zero_bias_ppl}
\begin{center}
\resizebox{0.8\linewidth}{!}{%
\begin{tabular}{ll cc r cc r}
\toprule
& & \multicolumn{3}{c}{PPL-CPT} & \multicolumn{3}{c}{PPL-Annealing} \\
\cmidrule(lr){3-5}\cmidrule(lr){6-8}
Method & Scenario & on & off & \multicolumn{1}{c}{$\Delta$} & on & off & \multicolumn{1}{c}{$\Delta$} \\
\midrule
Base    & --   & \multicolumn{2}{c}{2.40} & \multicolumn{1}{c}{--} & \multicolumn{2}{c}{5.46} & \multicolumn{1}{c}{--} \\
\midrule
Vanilla & cold & \multicolumn{2}{c}{2.19} & \multicolumn{1}{c}{--} & \multicolumn{2}{c}{7.23} & \multicolumn{1}{c}{--} \\
ALF-LB  & cold & 2.19 & 2.17 & $-0.9\%$ & 7.36 & 7.17 & $-2.6\%$ \\
PID (I) & cold & 2.19 & 2.17 & $-0.9\%$ & 7.39 & 7.20 & $-2.6\%$ \\
\midrule
Vanilla & warm & \multicolumn{2}{c}{2.24} & \multicolumn{1}{c}{--} & \multicolumn{2}{c}{6.07} & \multicolumn{1}{c}{--} \\
ALF-LB  & warm & 2.24 & 2.20 & $-1.8\%$ & 6.28 & 6.13 & $-2.4\%$ \\
PID (I) & warm & 2.23 & 2.19 & $-1.8\%$ & 6.28 & 6.14 & $-2.2\%$ \\
\bottomrule
\end{tabular}%
}
\end{center}
\end{table}

\paragraph{Conclusion.}
How much specialization a balancer costs depends on where it acts.
Aux-loss encodes the cost in the weights and pays a high price in all tested settings.
The bias-based balancers reach better balance at a smaller cost.
On a mixed corpus they cost almost nothing.
On a single-domain corpus, the deployed routing of ALF-LB and PID (I) loses about half of its routing information, and more than half of this loss returns once the bias is removed.
Removing the bias at inference sacrifices balance but lowers held-out perplexity, which indicates that the experts still work better with the gate's own selection than with the biased one.
We interpret this result as evidence that the bias reduces routing information and raises perplexity at deployment but leaves most of the input dependence learned by the gate weights intact.
We do not propose it as an inference procedure.
Training with the bias and discarding it at inference would trade inference-time balance for quality, and whether this trade holds at larger scale and in downstream evaluation remains open.

\subsection{Layer-wise metric}
\label{app:layerwise}

\begin{figure}[t]
\begin{center}
\includegraphics[width=0.9\textwidth]{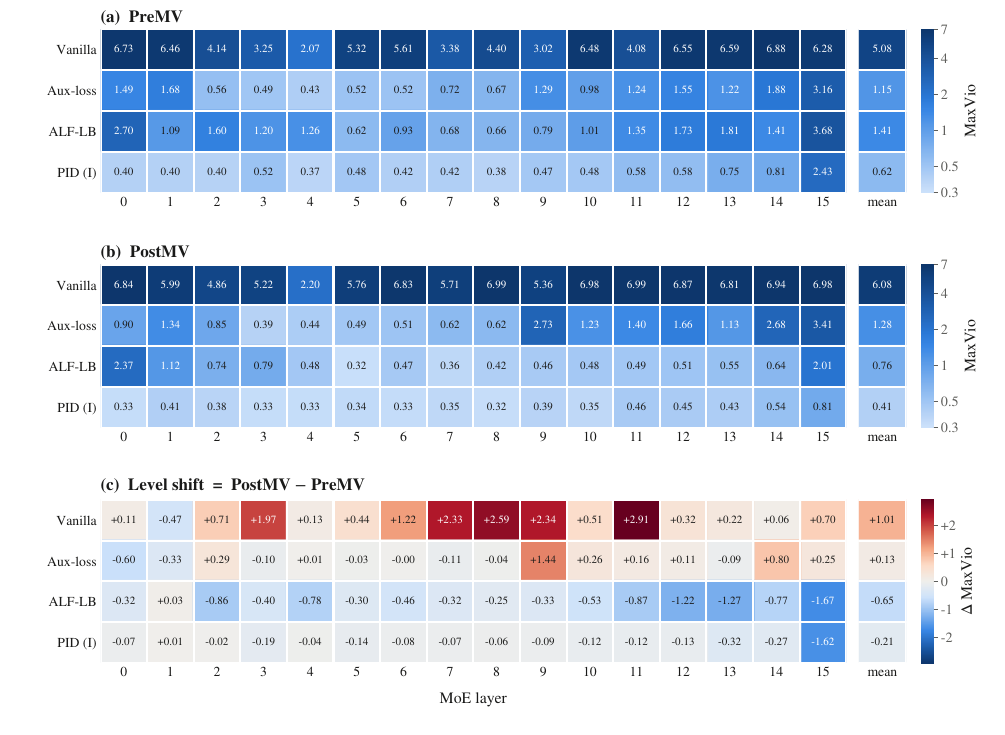}
\end{center}
\caption{Per-layer preMV (top), postMV (middle) and the level shift given by postMV minus preMV (bottom).}
\label{appfig:layerwise_maxvio}
\end{figure}

\begin{figure}[t]
\begin{center}
\includegraphics[width=0.9\textwidth]{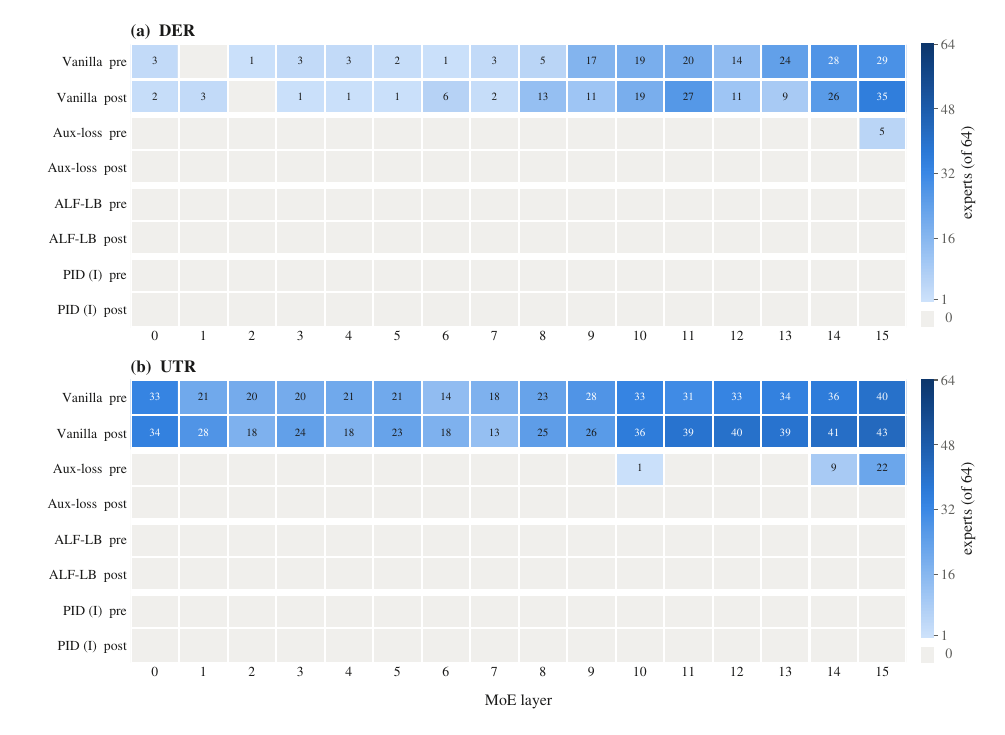}
\end{center}
\caption{Number of experts out of 64 in each layer whose window-mean share is below the dead (top) and under-trained (bottom) thresholds, in the pre and post windows.}
\label{appfig:layerwise_der}
\end{figure}

All balancing metrics in the main text are averaged over the sixteen MoE layers.
This section reports the same steady-state metrics per layer for the $E=64$, $1\to1$ pretraining cell as a representative example.
For demonstration, one seed per method is used here.
\Cref{appfig:layerwise_maxvio} shows preMV and postMV per layer, and \Cref{appfig:layerwise_der} shows the number of dead and under-trained experts per layer in the windows before and after the shift.

\paragraph{Steady-state imbalance.}
The spread across layers within one method is large.
Aux-loss and ALF-LB have a worst layer that is between roughly six and nine times more imbalanced than their best in both windows, and the last layer is the hardest or second hardest to balance for all three balancers.
ALF-LB additionally has a hard first layer, which becomes its worst layer after the shift.
Vanilla shows no such pattern because about half of its layers sit near the ceiling of the metric before the shift and most of them after it.
PID (I) has the lowest preMV and the lowest postMV in fifteen of the sixteen layers, and in the two remaining layers it is within 0.03 of the best baseline.
Its range across layers, taken as the worst layer minus the best, is also the narrowest of the four methods in both windows.
The level shift is negative in almost every layer for ALF-LB and PID (I) and positive in almost every layer for Vanilla, while Aux-loss has both signs.
This sign should not be read as the shift making balance easier or harder, since the two windows are taken from different corpora at different points in training.

\paragraph{Dead and under-trained experts.}
Vanilla has under-trained experts in every layer, with dead experts concentrated in the deeper half of the model and more than half of the experts under-trained in the deepest layers and in the first layer.
Aux-loss has dead or under-trained experts only in three of the deeper layers before the shift and none after it.
ALF-LB and PID (I) have no dead and no under-trained expert in any layer in either window.

\begin{figure}[t]
\begin{center}
\includegraphics[width=0.9\textwidth]{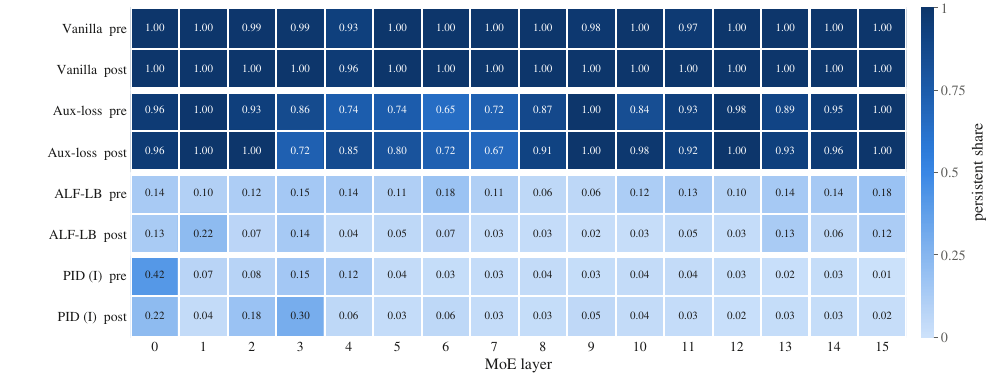}
\end{center}
\caption{Share of each layer's MaxVio carried by the expert that is most overloaded over the whole window.}
\label{appfig:layerwise_persist}
\end{figure}

\paragraph{Persistent versus rotating overload.}
Per-step MaxVio takes the most loaded expert at every step regardless of whether it is the same expert, whereas \Cref{appfig:layerwise_persist} accumulates the load over the whole window before taking the maximum and reports the ratio of this window-level MaxVio to the mean per-step MaxVio.
The ratio lies in $[0,1]$ and equals one only if the same expert is the most loaded at every step.
Vanilla stays above 0.93 in every layer and Aux-loss above 0.65, which means their imbalance persists on a single overloaded expert.
ALF-LB and PID (I) stay below 0.25 in all but two layers and never exceed 0.42, which means their per-step MaxVio measures a rotating overload rather than a permanently overloaded expert.

Taken together, the per-layer results show that the ranking of the four methods in the main text is not only driven by a few layers.

\subsection{Relating \texorpdfstring{$\text{MaxVio}$}{MaxVio} and \texorpdfstring{$\text{MRI}$}{MRI}}
\label{app:maxvio_mri}

\Cref{equ:MaxVio} and the maximum routing imbalance of \citet{therien2025continual} both normalize the load of the busiest expert.
This appendix derives the conversion used in \Cref{sec:discussion} and states the conditions under which it holds.

\paragraph{The two metrics.}
$\text{MaxVio}$ normalizes the busiest expert's load by the perfect balance target $r = TK/E$ of \Cref{equ:Error}, while $\text{MRI}$ normalizes the same load by the token count $T$,
\begin{equation}
\text{MaxVio}(t) = \frac{\max_i L_i(t) - r}{r},
\qquad
\text{MRI}(t) = \frac{\max_i L_i(t)}{T}.
\end{equation}
Only the denominators differ, and their ratio $T/r = E/K$ is a configuration constant.

\paragraph{Conversion.}
Factoring the common numerator $\max_i L_i(t)$ gives
\begin{equation}
\text{MaxVio}(t) + 1 \;=\; \frac{\max_i L_i(t)}{r} \;=\; \frac{\max_i L_i(t)}{T}\cdot\frac{T}{r} \;=\; \text{MRI}(t)\cdot\frac{E}{K},
\end{equation}
and therefore
\begin{equation}
\label{equ:mri_conversion}
\text{MRI}(t) = \bigl(\text{MaxVio}(t) + 1\bigr)\cdot\frac{K}{E},
\qquad
\text{MaxVio}(t) = \text{MRI}(t)\cdot\frac{E}{K} - 1 .
\end{equation}
The map is affine with positive slope $K/E$, hence either metric can be recovered from the other and their rankings agree once $E$ and $K$ are fixed.

\paragraph{Interpretation.}
The two normalizations answer two different questions about the same busiest expert.
$\text{MaxVio}$ asks how far that expert exceeds its target.
\Cref{equ:MaxVio} and \Cref{equ:Error} give $\text{MaxVio}(t) = -\min_i e_i(t)$, which is the largest per-expert balancer error at step $t$ and therefore the metric a feedback balancer is trying to drive to zero.
$\text{MRI}$ asks what fraction of all tokens in the batch land on that expert, which is the load its device must carry.
Under expert parallelism this share sets the compute of the slowest device, which is why \citet{therien2025continual} use it as a proxy for worst-case latency.

\paragraph{Endpoints.}
Under perfect balance every expert carries $r$, giving $\text{MaxVio} = 0$ and $\text{MRI} = K/E$.
Total collapse sends every token to the same group of $K$ experts, which then absorb the entire batch and give $\text{MRI} = 1$ together with the saturation ceiling $\text{MaxVio} = E/K - 1$.
$\text{MaxVio}$ thus fixes its zero at perfect balance and lets its ceiling vary with the configuration, whereas $\text{MRI}$ fixes its ceiling at $1$ and lets its floor vary.

\paragraph{Preconditions.}
The derivation requires two conditions, both of which hold in our experiments.
Routing must be dropless.
Every token then activates exactly $K$ experts and $\sum_i L_i(t) = TK$ holds at every step, which makes $r = TK/E$ a valid perfect-balance target.
Every token in the batch must also be routed rather than masked as padding, which keeps $T$ equal to the routed token count that both denominators rely on.

\paragraph{Example.}
Our main experiment presets fix $K/E = 0.125$.
Under the $7 \rightarrow 1$ shift at $E = 64$ in \Cref{tab:pretraining_results} the post-shift $\text{MaxVio}$ of Vanilla, Aux-loss, ALF-LB and PID (I) is $\{6.10, 1.50, 1.20, 0.47\}$.
\Cref{equ:mri_conversion} maps these values to a busiest-expert share of $\{0.89, 0.31, 0.28, 0.18\}$.
The excess over the balanced floor of $0.125$ is $\{0.76, 0.19, 0.15, 0.06\}$.
PID (I) leaves the smallest excess of the compared methods.

\subsection{Limitations and future work}
\label{app:limitation}
The largest MoE we train has 6.9B total and 1.3B active parameters.
Production MoEs are far larger in both total and active parameters.
At fixed routing sparsity the margin of our balancer over ALF-LB widens with expert count across our three scales, as \Cref{fig:maxvio_phase2} shows.
Three scales are too few to turn this trend into a firm conclusion, and whether it continues at production expert counts remains open.
Production training also consumes far more tokens over a broader corpus than the runs studied here.
Our runs therefore capture only part of the distribution shift a model meets over a full pretraining schedule.
Testing at that combined scale requires a training budget beyond what was available to us.

The distribution shift we study is a single hard switch between two domains, chosen to mimic the domain boundary of continual pretraining.
Gradual drift and repeated switches among more than two domains are not covered here.

Introducing the balancer requires modifying the router's forward pass, which is only possible when the full architecture and training code are open.
We selected OLMoE for this reason, and our results are therefore obtained on a single backbone.
Whether the advantages of our balancer transfer to other routing designs, such as expert-choice routing or routers with capacity limits, is left to future work.

The balancer we eventually adopt uses only the integral term, and within the present framework the error is the per-step deviation of each expert's load from its ideal share.
\Cref{app:pd} shows that adding the proportional or derivative term to this error, alone or together, does not improve on the integral term.
Both terms act on the instantaneous error or its change and are therefore exposed to the step-to-step fluctuation of the load, whereas the integral term averages that fluctuation out.
We do not rule out that a different error definition, for instance one smoothed over a window or built from a proxy of imbalance with smaller fluctuation, would let each term contribute its distinct advantage.
Finding such an error definition is the direction we consider most promising for the full PID family.

\end{document}